\documentclass[letterpaper]{article} 
\usepackage{aaai23}  
\usepackage{times}  
\usepackage{helvet}  
\usepackage{courier}  
\usepackage[hyphens]{url}  
\usepackage{graphicx} 
\usepackage{natbib}  
\usepackage{caption} 
\usepackage{algorithm}
\usepackage{algorithmic}

\usepackage{newfloat}
\usepackage{listings}
\DeclareCaptionStyle{ruled}{labelfont=normalfont,labelsep=colon,strut=off} 
\floatstyle{ruled}
\newfloat{listing}{tb}{lst}{}
\floatname{listing}{Listing}
\usepackage{mathtools}
\usepackage{amsmath, amssymb, amsthm} 
\usepackage{booktabs}
\usepackage[graphicx]{realboxes}
\usepackage{rotating,tabularx}
\usepackage{multirow}
\usepackage[utf8x]{inputenc} 
\usepackage{caption}
\usepackage{subcaption}

\newtheorem{theorem}{Theorem}
\newtheorem{proposition}[theorem]{Proposition}
\newtheorem{lemma}[theorem]{Lemma}

\newtheorem{definition}[theorem]{Definition}

\newcommand{\bfx}{\mathbf{x}}

\newcommand{\bfh}{\mathbf{h}}

\newcommand{\bff}{\boldsymbol{f}}
\newcommand{\bfg}{\boldsymbol{g}}
\newcommand{\bfp}{\boldsymbol{p}}
\newcommand{\bfw}{\mathbf{w}}
\newcommand{\bfW}{\mathbf{W}}
\newcommand{\bmth}{\boldsymbol{\theta}}

\newcommand{\bmdx}{\boldsymbol{\delta}_\bfx}
\newcommand{\hbmdx}{\boldsymbol{\hat{\delta}}_\bfx}

\newcommand{\calL}{\mathcal{L}}
\newcommand{\calD}{\mathcal{D}}
\newcommand{\bbR}{\mathbb{R}}
\newcommand{\nabx}{\nabla}
\newcommand{\hesg}{\mathbf{H}_{g}}
\newcommand{\hesf}{\mathbf{H}_{f}}

\title{Towards More Robust Interpretation via Local Gradient Alignment}

\author {
    Sunghwan Joo\textsuperscript{\rm 1*},
    Seokhyeon Jeong\textsuperscript{\rm 2*},
    Juyeon Heo\textsuperscript{\rm 3},
    Adrian Weller\textsuperscript{\rm 3,4},
    Taesup Moon \textsuperscript{\rm 2$\dagger$}
}
\affiliations {
    \textsuperscript{\rm 1} Department of Electrical and Computer Engineering, Sungkyunkwan University\\\
    \textsuperscript{\rm 2} Department of Electrical and Computer Engineering/ASRI/INMC/IPAI, Seoul National University\\\
    \textsuperscript{\rm 3} University of Cambridge\ \ \ 
    \textsuperscript{\rm 4} The Alan Turing Institute\\\
    \small{\texttt{shjoo840@gmail.com,sh102201@snu.ac.kr,\{jh2324,aw665\}@cam.ac.uk,tsmoon@snu.ac.kr}}
}

\begin{document}

\maketitle
\vspace{.05in}
\def\thefootnote{*}\footnotetext{Equal contribution. \ \ $ ^\dagger$Corresponding author.}

\begin{abstract}
Neural network interpretation methods, particularly feature attribution methods, are known to be fragile with respect to adversarial input perturbations. 
To address this, several methods for enhancing the local smoothness of the gradient while training have been proposed for attaining \textit{robust} feature attributions.
However, the lack of considering the normalization of the attributions, which is essential in their visualizations, has been an obstacle to understanding and improving the robustness of feature attribution methods. 
In this paper, we provide new insights by taking such normalization into account. First, we show that for every non-negative homogeneous neural network, a naive $\ell_2$-robust criterion for gradients is \textit{not} normalization invariant, which means that two functions with the same normalized gradient can have different values. 
Second, we formulate a normalization invariant cosine distance-based criterion and derive its upper bound, which gives insight for why simply minimizing the Hessian norm at the input, as has been done in previous work, is not sufficient for attaining robust feature attribution. Finally, we propose to combine both $\ell_2$ and cosine distance-based criteria as regularization terms to leverage the advantages of both in aligning the local gradient. As a result, we experimentally show that models trained with our method produce much more robust interpretations on CIFAR-10 and ImageNet-100 without significantly hurting the accuracy, compared to the recent baselines. To the best of our knowledge, this is the first work to verify the robustness of interpretation on a larger-scale dataset beyond CIFAR-10, thanks to the computational efficiency of our method.
\end{abstract}

\section{Introduction}
\label{sec:introduction}

Feature attribution methods \cite{simonyan2014deep,shrikumar2017learning,springenberg2014striving,bach2015pixel}, which refer to interpretation methods that numerically score the contribution of each input feature for a model output, have been 
useful tools to reveal the behavior of complex models, \textit{e.g.}, deep neural networks, especially in application domains that require safety, transparency, and reliability.
However, several recent works \cite{ghorbani2019interpretation, Dombrowski2019ExplanationsCB, kindermans2019reliability} identified that such methods are vulnerable to adversarial input manipulation, namely, 
an adversarially chosen imperceptible input perturbation can arbitrarily change feature attributions without hurting the prediction accuracy of the model. One plausible explanation for this vulnerability can be made from a geometric perspective \cite{Dombrowski2019ExplanationsCB}, namely, if the decision boundary of a model is far from being smooth, \textit{e.g.,} as in typical neural networks with ReLU activations, a small movement in the input space can dramatically change the direction of the input gradient, which is highly correlated with modern feature attribution methods. 


To defend against such adversarial input manipulation, 
recent approaches \cite{wang2020smoothed,dombrowski2022towards} regularized the neural network to have locally smooth input gradients while training. 
A popular criterion to measure the smoothness of the input gradients is to use the $\ell_2$-distance between the gradients of the original and perturbed input points, dubbed as the $\ell_2$ robust criterion. Since this $\ell_2$-criterion can be upper bounded by the norm of the Hessian with respect to the input, \cite{wang2020smoothed,dombrowski2022towards} used the \textit{approximated} norm of the Hessian as a regularization term while training. Furthermore, \cite{dombrowski2022towards} shows that replacing the ReLU activation with the Softplus function, Softplus$_{\beta}(x) = \frac{1}{\beta}\log(1+\exp(\beta x))$, can 
smooth the decision boundary and lead to a more robust feature attribution method. 

\begin{figure}[t]
\vskip 0.2in
\begin{center}
\centerline{\includegraphics[width=0.8\columnwidth]{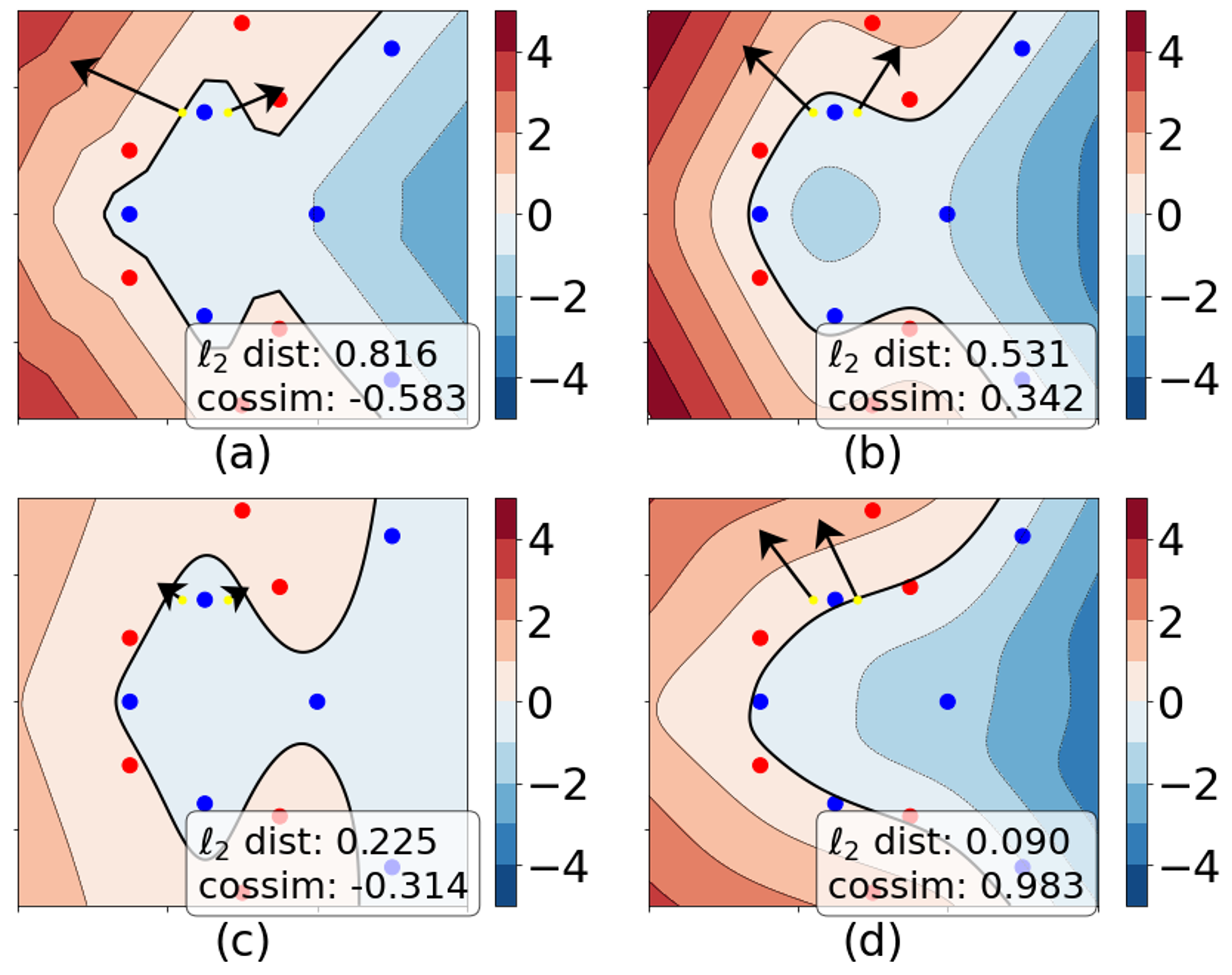}}
\caption{Visualization of the level curves and gradients (shown in arrows) of two nearby input points near the decision boundary of two-layer neural networks, which are trained on a synthetic binary labeled dataset. The $\ell_2$ distance and cosine similarities of the two gradient vectors of logit are shown in the legend box. Each figure corresponds to a network trained with
\textbf{(a)} ReLU activation with Binary Cross Entropy (BCE) loss. 
\textbf{(b)} Softplus activation with BCE loss.
\textbf{(c)} Softplus activation with BCE loss and a MaxEnt regularizer.
\textbf{(d)} Softplus activation with BCE loss and our proposed robust criterion.
}
\vspace{-.1in}
\label{fig:decision_surface}
\end{center}
\vskip -0.2in
\end{figure}

While the above approach was shown to be effective to some extent, we argue that only naively considering the $\ell_2$ robust criterion is limited since it does not take the normalization of the attributions into account. Such normalization is essential in visualizing the attributions, since the \textit{relative} difference between the attributions is most useful and important for interpretation.
To that end, we argue that simply trying to reduce the $\ell_2$ robust criterion or the norm of the Hessian with respect to the input while training may not always lead to robust attributions. As concrete examples, consider the level curves and gradients of four differently trained two-layer neural networks in Figure \ref{fig:decision_surface}. In Figure \ref{fig:decision_surface}(a), we clearly observe that the kinks (continuous but not differentiable points) generated by the ReLU activation can cause a dramatic change in the directions of the gradients of two nearby points. In contrast, as shown in Figure \ref{fig:decision_surface}(b), the Softplus activation smoothes the decision surface, resulting in a less dramatic change of the gradients compared to Figure \ref{fig:decision_surface}(a) (\textit{i.e.,} smaller $\ell_2$ distance and larger cosine similarity). An interesting case is Figure \ref{fig:decision_surface}(c) that uses the Maximum Entropy (MaxEnt) regularization, which is known to promote a wide-local minimum \cite{cha2020cpr} and a small Hessian norm at the trained model parameter, while training.  We observe the $\ell_2$ distance between the two local gradients has certainly shrunk due to the small norm of the Hessian at the input, but the cosine similarity between them is very low as well; namely, the decrease of the $\ell_2$ distance is mainly due to the decrease of the norm of the gradients. Thus, after the normalization of the gradients, the two local gradients still remain very different, leaving the fragility of the interpretation unsolved. This example shows that naively minimizing the Hessian norm as a proxy for robust attribution may not \textit{necessarily} robustify the attribution methods. An ideal case would be Figure \ref{fig:decision_surface}(d) in which the $\ell_2$ distance is reduced by also \textit{aligning} the local gradients as we propose in the paper.

In this paper, we propose to develop a more robust feature attribution method, 
by promoting the alignment of local gradients, hence making the attribution more invariant with respect to the normalization. 
%
%
%
%
%
%
More specifically, we first define a \textit{normalization invariant criterion}, which refers to the criterion that has the same value for two functions with the same normalized input gradient. 
Then, we show that the $\ell_2$ robust criterion is \textit{not} normalization invariant by leveraging the non-negative homogeneous property of the ReLU network. We then suggest considering cosine distance-based  criterion and show that it \textit{is} normalization invariant and is upper bounded by the ratio of the norm of the Hessian and the norm of the gradient at the input. Our theoretical finding explains why simply minimizing the Hessian norm may not necessarily lead to a robust interpretation as in the example of Figure \ref{fig:decision_surface}(c).
Finally, we propose to combine both the $\ell_2$ and cosine robust criteria as regularizers while training and show on several datasets and models that our method can achieve promising interpretation robustness. 
Namely, our method has better robustness with respect to several quantitative metrics compared to other recent baselines and the qualitative visualization for adversarial input also confirmed the robustness of our method. Furthermore, we stress that the computational efficiency of our method enabled the first evaluation of the interpretation robustness on a larger-scale dataset, \textit{i.e., }ImageNet-100, beyond CIFAR-10.

%

\section{Related Works}

\subsubsection{Vulnerability of attribution methods}
Attention toward trustworthy attributions has developed several recent works showing that the attribution methods 
are susceptible to adversarial or even random perturbations on input or model. \cite{adebayo2018sanity, kindermans2019reliability, sixt2020explanations} show that some popular attribution methods generate visually plausible attributes, namely, they are independent of the model or too sensitive to a constant shift to the input. Inspired by adversarial attacks \cite{goodfellow2015explaining,madry2018towards}, \cite{ghorbani2019interpretation, Dombrowski2019ExplanationsCB} identify that malign manipulation of attributions can be produced by an imperceptible adversarial input perturbation with the same prediction of the model. On the other hand, \cite{heo2019fooling, slack2020fooling, lakkaraju2020fool,dimanov2020you} changes the model parameters to manipulate the attributions while keeping the model output unchanged. 

\subsubsection{Towards attribution robustness}
Recent efforts toward developing robust attributions have been made from various perspectives. \cite{Laura2020simpledefense} suggested that simply aggregating multiple explanation methods can enhance the robustness of attribution. \cite{pan2020fairwashing} proposed a projection-based method that eliminates the off-manifold components in attribution which mainly cause vulnerable attributions. \cite{smilkov2017smoothgrad, si2021simple} proposed aggregation of explanations over randomly perturbed inputs.
The above works have a different focus than ours, since they do not consider model training methodology for attaining robust attributions. 
Minimization over worst-case input is another well-known strategy for attaining robust attribution. 
\cite{chen2019robust, ivankay2020far} suggested adversarial attributional training for an $\ell_p$ norm of Integrated Gradient \cite{sundararajan2017axiomatic}.
Likewise, \cite{singh2020attributional} proposed a min-max optimization with soft-margin triplet loss to align input and gradients. However, the inner-maximization process is computationally heavy, which is the main obstacle to applying them in large-scale datasets. 

\subsubsection{Smoothing the geometry}
Smoothing the geometry of the model is a popular approach to building both robust attributions and robust classification. 
For robust attribution, 
\cite{dombrowski2022towards, wang2020smoothed} approximated the norm of the Hessian matrix and regularized it to give a penalty on the principal curvatures. \cite{tang2022defense} proposed knowledge distillation that sample the input by considering the geometry of the teacher model. 
For adversarial robustness, 
\cite{qin2019adversarial} encourages the loss to behave linearly, and \cite{andriushchenko2020understanding} reduces cosine distance between gradients of closely located points to prevent catastrophic overfitting that conventionally happens during adversarial training. \cite{singla2021low, jastrzebski2021catastrophic} analyze how smoothing the curvature can increase adversarial robustness and generalization. Compared to them, our method is the first to apply cosine distance for attaining robust attribution via promoting smooth geometry of the neural networks, with reduced computational complexity. 

\section{Preliminaries}
\subsection{Notation}
We consider a set of data tuples $\calD = \{(\bfx^{(n)}, y^{(n)})\}_{n=1,...,N}$, where $\bfx^{(n)} \in \bbR^d$ is an input data and $y^{(n)} \in \{1,2,...,C\}$ is the target label. 
We denote a neural network classifier as $\bfg: \bbR^d \rightarrow \bbR^C$ which returns a logit value for each class.
Also, we denote a function $\bfp = \sigma_\text{softmax} \circ \bfg : \bbR^{d} \rightarrow \vartriangle^C$ as a composition of the logit and the softmax function that returns a categorical probability, where $\vartriangle^C$ denotes a $C$-dimensional simplex. We denote an attribution function by $\bfh: \bbR^{d} \rightarrow \bbR^{d}$, where each element value of $\bfh$ represents the importance or sensitivity of the input for the output. We always use a bold symbol for vector and vector-valued functions, \textit{e.g.}, $\bfx$, and $\bfg$, respectively. 
We use a subscript indexing, \textit{e.g.,} $x_i$ is the $i$-th element of a vector $\bfx$ and $g_c$ is a scalar-valued function which returns the $c$-th element of $\bfg(\bfx)$. We denote a scalar-valued function by $f: \bbR^d \rightarrow \bbR$ and denote a gradient of $f$ with respect to $\bfx$ by $\nabx f(\bfx) = (\frac{\partial}{\partial x_1} f(\bfx), \frac{\partial}{\partial x_2} f(\bfx), ...,\frac{\partial}{\partial x_d} f(\bfx))$.
Also, we denote a Hessian matrix of function $f$ with respect to $\bfx$ by $\mathbf{H}_{f}(\bfx)$, where $(\mathbf{H}_{f}(\bfx))_{ij} = \frac{\partial^2}{\partial x_i \partial x_j} f(\bfx)$. If there is no ambiguity, we omit the symbol $(n)$ from $(\bfx^{(n)}, y^{(n)})$.

\subsection{Adversarial attribution manipulation}
\label{sec:adv_attack_and_robustness}
We first introduce Adversarial Attribution Manipulation (AAM) proposed in \cite{ghorbani2019interpretation,Dombrowski2019ExplanationsCB}. The aim of AAM is to find an adversarial example $\bfx_{\text{adv}}$ in a given $\ell_p$ ball around $\bfx$, $||\bfx_{adv} - \bfx||_p \leq \epsilon$, such that the attribution changes significantly, while the prediction remains unchanged. The AAM is categorized as \textit{targeted}, if the goal of the adversary is to make the attribution similar to a target map. In contrast, the AAM is called \textit{untargeted}, if the adversary makes the attribution as dissimilar as possible compared to the original attribution. The objective of AAM is formally given by
\begin{align*}
    &\text{minimize}_{||\bmdx||_p \leq \epsilon} \Lambda_\bfh(\bfx, \bmdx) \\
    &\text{s.t. } \text{argmax}_c g_c(\bfx + \bmdx) = \text{argmax}_c g_c(\bfx),
\end{align*}
in which $\bmdx \triangleq \bfx_{adv} - \bfx$ is an input perturbation and $\Lambda_\bfh(\bfx, \bmdx)$ is the attacker's criterion. In this paper, we mainly consider a targeted AAM proposed in \cite{Dombrowski2019ExplanationsCB}, in which $\Lambda_\bfh(\bfx, \bmdx) = ||\bfh(\bfx + \bmdx) - \bfh_t||_2$ is used with $\bfh_t$ being a target map.

To defend against  AAM, previous works \cite{wang2020smoothed,dombrowski2022towards} considered minimizing an $\ell_2$ robust criterion which is the $\ell_2$ difference between the gradients of logit with respect to closely located points. 
Formally, the $\ell_2$ robust criterion of a function $\bfg$ at a given tuple  $(\bfx, y)$ is denoted and defined by 
\begin{align}
\Gamma_{\nabx g_y}^{\ell_2}(\bfx, \bmdx) \coloneqq ||\nabx g_y(\bfx + \bmdx) - \nabx g_y(\bfx)||_2,\nonumber 
\end{align}
where $\bmdx$ is a perturbation.
From this definition, we re-write the criterions and their upper bounds that were proposed in \cite{dombrowski2022towards} and \cite{wang2020smoothed} as
\begin{align}
&\Gamma_{\nabx g_y}^{\ell_2}(\bfx, \bmdx) \leq ||\mathbf{H}_{g_y}(\bfx)||_F L(\bfx_{adv}, \bfx) \nonumber \\
&\text{max}_{||\bmdx||_p\leq \epsilon} \Gamma_{\nabx \mathcal{L}}^{\ell_2}(\bfx, \bmdx) \leq \epsilon ||\mathbf{H}_\mathcal{L}(\bfx)||_2,
\label{eq:hessian_upper_bound}
\end{align}
respectively, in which $L(\bfx_{adv}, \bfx)$ is a distance between $\bfx_{adv}$ to $\bfx$ and $\mathcal{L}$ denotes a cross-entropy loss. Then, \cite{wang2020smoothed,dombrowski2022towards} approximated the Hessian and used its norm as a regularizer during training to minimize the upper bound on the $\ell_2$ robust criterion.


\noindent\textit{Remarks 1:} Note that the obtained attributions are conventionally normalized, \textit{i.e.}, in case of the vision domain, attribution values are mapped into the range $\{0, ..., 255\}$ for visualization. Also, the attacker usually normalizes both $\bfh(\bfx + \bmdx)$ and $\bfh_t$ to match the scale, because $\bfh_t\in\{0, 1\}^d$ in usual, while $\bfh(\bfx + \bmdx)$ has a much diverse range.\\
\noindent\textit{Remarks 2:} If a neural network is a piece-wise linear function, the Hessian matrix of logit $\mathbf{H}_{g_c}(\bfx)$ is a zero matrix. 
To make Hessian non-zero, \cite{wang2020smoothed, moosavi2019robustness} considered the Hessian of the cross entropy loss, instead of the logit function. However, this alternative would still contain non-differentiable points. On the other hand, \cite{dombrowski2022towards} introduced a Dirac delta function to deal with a non-differentiable point, i.e., ReLU$''(x)=\delta(x)$.
Besides, \cite{ghorbani2019interpretation} considered replacing the ReLU with Softplus, which is twice differentiable and well approximates ReLU.

\section{Theoretical Consideration}

\subsection{Normalization invariant criterion}
\label{subsec:normalization_invariant}

As we described in Figure 1, the reduction of the $\ell_2$ robust criterion or Hessian norm may not always lead a model to be robust against the AAM, because such reduction can be also achieved by decreasing the norms of the gradients without actually aligning them. 
Such limitation of the $\ell_2$ criterion motivates us to devise a criterion that would not be affected by the normalization of the gradient. We formally define a \textit{normalization invariant criterion} as follows.
\begin{definition}
\label{def:normalization_invariant}
For given functions $\bff:\bbR^d \rightarrow \bbR^C$ and $\bfg:\bbR^d \rightarrow \bbR^C$ that satisfy $\nabx f_y(\bfx) / ||\nabx f_y(\bfx)||_2 = \nabx g_y(\bfx) / ||\nabx g_y(\bfx)||_2$ for all $(\bfx, y) \in \bbR^d \times \{1,2,...,C\}$, a criterion $\Gamma$ is called \textbf{normalization invariant} (for $\bff$ \& $\bfg$), if 
\begin{align*}
    \Gamma_{\nabx f_y}(\bfx, \bmdx) = \Gamma_{\nabx g_y}(\bfx, \bmdx),
\end{align*}
in which $\Gamma_{f}(\bfx, \bmdx)$ stands for some function that measures the distance between $f(\bfx)$ and $f(\bfx+\bmdx)$.
\end{definition}

Next, we show that the $\ell_2$ robust criterion $\Gamma_{\nabx f}^{\ell_2}$ is not normalization invariant for any neural network of which the final layer is a fully connected layer. 
Without loss of generality, we denote a set of parameters for a neural network as $\Theta = \{\bmth_1, \bmth_2, ..., \bmth_L\}$, where each $\bmth_i$ refers to the parameter of the $i$-th layer. We define an $\alpha$-transformation as $T_i(\Theta, \alpha) = \{ \bmth_1, ..., \bmth_{i-1}, \alpha\bmth_i, \bmth_{i+1}, ..., \bmth_L \}$ which scales the $i$-th parameter by $\alpha>0$.
Then, we define a notion of \textit{non-negative homogeneous} function as follows. 
\begin{definition}
\label{def:non_negative_homogeneous}
A function $\bff:\mathbb{R}^d \to \mathbb{R}^C$ with 
parameters $\Theta = \{\bmth_1, \bmth_2, ..., \bmth_L\}$ is \textbf{non-negative homogeneous} with respect to parameter $\bmth_i$, if $\bff$ satisfies the following for all $\alpha \in \mathbb{R_+}$ and $\bfx \in \bbR^{d}$:
\begin{align*}
    \bff(\bfx;T_i(\Theta, \alpha))=\alpha \bff(\bfx;\Theta).
\end{align*}
\end{definition}

We note that every neural network with the final fully-connected layer is non-negative homogeneous with respect to the parameters of the last layer. Moreover, a feed-forward neural network with ReLU activation can be shown to be non-negative homogeneous, with respect to \textit{any} parameters, of which proofs are available in Appendix B. Now, we show the following lemma on the $\ell_2$ robust criterion for non-negative homogeneous neural networks. 

\begin{lemma}
\label{lemma:crc_is_nnh}
The $\ell_2$ robust criterion is not normalization invariant for non-negative homogeneous neural networks. 
\end{lemma}
\noindent\emph{Proof:} Let $\bfg$ be the non-negative homogeneous neural network. Then, for any output class $c$, by performing the $\alpha$-transformation, we can set $f(\bfx)=\alpha g_y(\bfx)$ for any $\alpha\in\mathbb{R}_{+}$ and $\bfx\in\mathbb{R}^d$. Then, it is clear that
$
    \nabx g_y(\bfx) / ||\nabx g_y(\bfx)||_2 = \alpha\nabx g_y(\bfx) / ||\alpha\nabx g_y(\bfx)||_2 = \nabx f(\bfx) / ||\nabx f(\bfx)||_2,
$  
but we have
\begin{align}
    \Gamma_{\nabx f}^{\ell_2}(\bfx, \bmdx) &= ||\nabx f(\bfx + \bmdx) - \nabx f(\bfx)||_2 \nonumber\\
    &= \alpha||\nabx g_y(\bfx + \bmdx) - \nabx g_y(\bfx)||_2 \nonumber\\
    &=\alpha\cdot \Gamma_{\nabx g_y}^{\ell_2}(\bfx, \bmdx).\label{eq:nonnegative}
\end{align}
Therefore, $f$ and $g_y$ can have different $\ell_2$ criterion values although they have the same normalized gradient. \qedsymbol

The above lemma implies that minimizing the $\ell_2$ robust criterion for non-negative homogeneous neural networks may not be satisfactory for obtaining robust interpretations. The reason is that it may simply result in scaling the network parameters such that the criterion is minimized, but the argmax prediction and the gradient direction for the network remain unchanged. 


\subsection{A cosine robust criterion}
\label{subsec:ojb_and_upperbound}

From the result of the previous section, we
propose \textit{cosine robust criterion} (CRC) which 
measures the cosine distance between the gradients at two nearby points.
\begin{definition}\label{def:crc}
A cosine robust criterion of a function $\bfg$ at point $(\bfx, y)$ is denoted and defined by
\begin{equation*}
    \Gamma_{\nabx g_y}^{\text{\emph{cos}}}(\bfx, \bmdx) \coloneqq \frac{1}{2}\big(1-\text{cossim}\left(\nabx g_y(\bfx + \bmdx), \nabx g_y(\bfx)\right)\big), \label{eq:crc}
\end{equation*}
in which $\text{cossim}(\mathbf{v}, \mathbf{w}) = \mathbf{v}^T \mathbf{w} / (||\mathbf{v}||_2 \cdot || \mathbf{w}||_2)$.
\end{definition}

\begin{proposition}
\label{prop:ni_cosine_distance}
The cosine robust criterion is normalization invariant for any $\bff$ and $\bfg$ that satisfies $\nabx f_y(\bfx) / ||\nabx f_y(\bfx)||_p = \nabx g_y(\bfx) / ||\nabx g_y(\bfx)||_p$ for all $(\bfx, y)$.
\end{proposition}

The proof of  Proposition \ref{prop:ni_cosine_distance} is trivial because the normalization is self-contained in the cosine similarity. We now attain the upper bound on CRC as follows. 

\begin{theorem}
\label{thm:upper_bound}
For a twice differentiable function $g$, $\epsilon>0$ with $\epsilon \ll 1$, and $\forall (\bfx, \bmdx) \in \bbR^d\times\bbR^d$, we have 
\begin{align}
    \Gamma_{\nabx g}^{\text{\emph{cos}}}(\bfx, \epsilon\hbmdx)
    \leq \frac{\epsilon \|\hesg(\bfx)\|_F  + \mathcal{O}(\epsilon^2)}
    {\|\nabx g(\bfx+\epsilon\hbmdx)\|_2},\label{eq:upper bound}
\end{align}
in which 
$\hbmdx = \bmdx/||\bmdx||_2$.
\end{theorem}
\noindent\textit{Proof:} See Appendix B. \qedsymbol

In the theorem, without loss of generality, we substituted $\bmdx$ in the CRC with a multiplication of scalar and unit vector, denoted as $\epsilon\hbmdx$, and we
applied the Taylor expansion $\nabx g(\bfx+\epsilon\hbmdx) = \nabx g(\bfx) + \epsilon\hesg(\bfx)\hbmdx + \mathcal{O}(\epsilon^2)\mathbf{v}$. 
This theorem implies that only minimizing the Hessian norm as in \cite{wang2020smoothed,dombrowski2022towards} may not be sufficient to align the local gradients, \textit{i.e.,} minimize the CRC.
As a simple example, consider again a non-negative homogeneous neural network $\bfg$ that can be $\alpha$-transformed into  $f(\bfx)=\alpha g_y(\bfx)$ for any output class $y$ as in the proof of Lemma \ref{lemma:crc_is_nnh}. Then, we observe that the Hessian norm of $f$ can be easily minimized since $\|\hesf(\bfx)\|_2=\alpha \|\mathbf{H}_{g_y}(\bfg)\|_2$ and $\alpha$ can be made arbitrarily small. However, since we also have $\|\nabx f(\bfx)\|_2=\alpha\|\nabx g_y(\bfx)\|_2$ for any $\bfx$, the upper bound in (\ref{eq:upper bound}) would not shrink for $f$, showing that a simple $\alpha$-transformation of a non-negative homogeneous neural network would not necessarily minimize CRC as opposed to the $\ell_2$ robust criterion in Lemma \ref{lemma:crc_is_nnh}.



\subsection{Proposing Methods}
\label{subsec:regularization}


From the theoretical considerations of the previous section, we may first propose a straightforward way to directly apply the CRC to training. Namely,
for a single data tuple $(\bfx, y)$ and model $\bfg$, we denote a training loss function with regularization as $\calL(\bfx, y)$ and define it as
\begin{align}
    \calL(\bfx, y) = \calL_{CE}(\bfx, y) + \lambda_{\text{cos}} \mathbb{E}_{\bmdx} \Big[\Gamma_{\nabx g_y}^{\text{cos}}(\bfx, \bmdx)\Big],
\end{align}
in which $\calL_{CE}(\bfx, y)$ is the cross-entropy loss, $\bmdx$ is sampled from $\mathcal{U}_d ([-\epsilon, \epsilon]^d)$, a $d$-dimensional multivariate uniform distribution on an $\ell_\infty$ $\epsilon$-ball, and $\lambda_{\text{cos}}$ is a hyper-parameter for determining the CRC regularization strength.

While using CRC as a regularizer seems promising in the sense of maintaining similar directions for local gradients, we argue that only considering the angle between local gradients by CRC might cause instability while training and large variability of the magnitude of the gradients. 
An extreme case that could occur in training is that we enforce a similar derivative direction in a ball around $\bfx$, but as we move in some direction from $\bfx$ towards the edge of the ball, the magnitude of the derivative could become very low. In this case, if we were to continue in the same direction just beyond the edge of the ball, the direction of the derivative could flip and point in the opposite direction. This phenomenon may reduce training stability and generalization of behavior beyond the training points.  
%
%

With the above reasoning, our final proposal is to combine both cosine and $\ell_2$ robust criteria as regularizers:
\begin{align}
    &\calL(\bfx, y) = \calL_{CE}(\bfx, y) \nonumber \\
    & + \mathbb{E}_{\bmdx } \Big[ \lambda_{\text{cos}} \Gamma_{\nabx g_y}^{\text{cos}}(\bfx, \bmdx) + \lambda_{\ell_2} \Gamma_{\nabx g_y}^{\ell_2}(\bfx, \bmdx)\Big],
    \label{eq:crcr}
\end{align}
in which $\lambda_{\ell_2}$ is a hyper-parameter for the $\ell_2$ regularization strength, and $\bmdx$ is again sampled from $\mathcal{U}_d ([-\epsilon, \epsilon]^d)$.
Both regularizers complement each other: the CRC helps to align the local gradients, while the $\ell_2$ regularizer contributes to stable training steps. 
For every training iteration, we do a Monte-Carlo sampling for each data point to sample $\bmdx$. 
While we do not consider the worst-case perturbation as in adversarial training \cite{madry2018towards}, we instead try to align the local gradients within the $\ell_\infty$ $\epsilon$-ball to achieve probable robustness.






\section{Experiments}

\subsection{Attribution methods}
We measure the robustness of attribution methods that are known to be closely related to the input gradient, \textit{i.e.}, Gradient, Input$\times$Gradient, Guided Backprop, and LRP. These methods generate saliency maps, showing which pixels are most relevant to the classifier output. 
We denote the saliency map for model $\bfg$ at data tuple $(\bfx, y)$ using attribution method $\mathcal{I}$ as $\bfh_{g_y}^\mathcal{I}(\bfx)$.

\noindent\textbf{Gradient} \cite{simonyan2014deep}: This method is 
defined by $\bfh_{g_y}^{\text{Grad}}(\bfx) = \nabx g_y(\bfx)$, which is the gradient of a logit with respect to input. 

\noindent\textbf{Input$\times$Gradient} \cite{shrikumar2017learning}: This method calculates the element-wise multiplication of input and gradient,  $\bfh_{g_y}^{\times\text{Grad}}(\bfx) = \bfx \odot \nabx g_y(\bfx)$.

\noindent\textbf{Guided Backprop} \cite{springenberg2014striving}: This method is a modified back-propagation method that does not propagate the true gradient, but the imputed gradient when it passes through the activation function. Specifically, if the forward pass of the activation is given by $a_i = \text{max}(z_i, 0)$, the modified propagation is represented as $\nabla_{z_i}g_y = \nabla_{a_i}g_y \cdot \mathbf{1}(\nabla_{a_i} g_y>0) \cdot \mathbf{1}(z_i>0)$. We denote this method as $\bfh_{g_y}^{GBP}$.

\noindent\textbf{LRP} \cite{bach2015pixel}: LRP 
propagates relevance from output to input layer-by-layer, based on the contributions of each layer. 
We used the z+ box rule in our experiments, as adopted in \cite{dombrowski2022towards}. Starting from the relevance score of the output layer $R_i^{(L)} = \delta_{iy}$, we apply the $z^+$ propagation rule for the intermediate layers
\begin{align*}
    R_i^{(\ell)} = \sum_j \frac{x_i^{(\ell)}(W_{ij}^{(\ell)})_+}{\sum_{i'}x_{i'}^{(\ell)}(W_{i'j}^{(\ell)})_+} R_j^{(\ell+1)},
\end{align*}
where $\bfW^{(\ell)}$ and $x_i^{(\ell)}$ denote the weights and activation vector of the $\ell$-th layer, respectively. For the last layer, we use the $z^\mathcal{B}$ rule to bound the relevance score in the input domain,
\begin{align*}
    R_i^{(0)} = \sum_j \frac{x_i^{(0)}W_{ij}^{(0)} - l_i (W_{ij}^{(0)})_+ - h_i (W_{ij}^{(0)})_-}{\sum_{i'} x_{i'}^{(0)}W_{i'j}^{(0)} - l_{i'} (W_{i'j}^{(0)})_+ - h_{i'} (W_{i'j}^{(0)})_-} R_j^{(1)},
\end{align*}
where $l_i$ and $h_i$ are the lower and upper bounds of the input domain. The heatmap is then denoted as $\bfh^{LRP}_{g_y}(\bfx) = \mathbf{R}^{(0)}$.

\begin{table*}[ht]
\centering
\caption{Quantitative results}
\label{tab:quantitative_sumary}
\resizebox{\textwidth}{!}{
\begin{tabular}{c|c|c|ccc|ccc|ccc|ccc}
\toprule
\multirow{2}{*}{\shortstack{Dataset\\Model}} & \multirow{2}{*}{Regularizer} & \multirow{2}{*}{Accuracy} &  \multicolumn{3}{c|}{Grad \shortcite{simonyan2014deep}} & \multicolumn{3}{c|}{xGrad \shortcite{shrikumar2017learning}} & \multicolumn{3}{c|}{GBP \shortcite{springenberg2014striving}} & \multicolumn{3}{c}{LRP \shortcite{bach2015pixel}} \\
             &     &  &           RPS & Ins & A-Ins &             RPS & Ins & A-Ins &              RPS & Ins & A-Ins &         RPS & Ins & A-Ins \\
\midrule
\multirow{6}{*}{\shortstack{CIFAR10\\LeNet}} & CE only & 86.8 &           0.760 &  35.5 &  29.1 &             0.749 &  43.1 &  37.4 &            0.897 &  51.0 &  43.0 &         0.980 &  56.6 &  48.4 \\
             & ATEX \cite{tang2022defense} & 88.0 &           0.771 &  35.0 &  28.9 &             0.760 &  43.0 &  37.5 &            0.907 &  51.5 &  42.7 &         0.982 &  58.0 &  49.9 \\
             & IGA \cite{singh2020attributional} & 86.3 &                             0.837 &  37.9 &  34.0 &                                   0.822 &  47.2 &  44.6 &                                    0.937 &  57.7 &  54.7 &                         0.994 &  59.1 &  55.1 \\
             & Hessian \cite{dombrowski2022towards} & 86.3 &           0.872 &  37.6 &  31.7 &             0.861 &  45.8 &  41.6 &            0.960 &  59.4 &  54.8 &         \textbf{0.993} &  59.2 &  56.3 \\
             & $\ell_2$ & 86.4 &           0.881 &  41.1 &  33.4 &             0.873 &  48.2 &  42.3 &            0.963 &  59.5 &  55.4 &         0.992 &  59.9 &  56.5 \\
             & $\ell_2$+ Cosd (ours) & 86.1 &           \textbf{0.895} &  \textbf{43.3} &  \textbf{35.8} &             \textbf{0.887} &  \textbf{50.5} &  \textbf{44.8} &            \textbf{0.966} &  \textbf{61.4} &  \textbf{57.4} &         0.992 &  \textbf{60.3} &  \textbf{57.1} \\
\midrule
\multirow{4}{*}{\shortstack{CIFAR10\\ResNet18}} & CE only & 94.0 &           0.724 &  40.8 &  31.7 &             0.708 &  47.7 &  39.6 &            0.902 &  60.1 &  53.8 &         0.972 &  64.5 &  59.8 \\
             & Hessian \cite{dombrowski2022towards} & 93.3 &           0.864 &  46.6 &  38.3 &             0.852 &  53.9 &  48.0 &            0.961 &  62.3 &  57.7 &         0.988 &  70.3 &  68.3 \\
             & $\ell_2$ & 93.3 &           0.908 &  49.3 &  43.3 &             0.898 &  56.7 &  52.2 &            \textbf{0.976} &  65.9 &  \textbf{65.1} &         0.988 &  \textbf{71.1} &  \textbf{70.3} \\
             & $\ell_2$+ Cosd (ours) & 93.0 &           \textbf{0.931} &  \textbf{53.4} &  \textbf{48.9} &             \textbf{0.924} &  \textbf{58.7} &  \textbf{55.3} &            0.974 &  \textbf{66.1} &  \textbf{65.1} &         \textbf{0.989} &  70.8 &  69.7 \\
\midrule
\multirow{4}{*}{\shortstack{ImageNet100\\ResNet18}} & CE only & 79.0 &           0.790 &  39.8 &  29.8 &             0.777 &  44.7 &  34.1 &            0.909 &  \textbf{49.1} &  34.0 &         0.948 &  50.1 &  35.3 \\
           & Hessian \cite{dombrowski2022towards} & 78.9 &                             0.830 &  42.4 &  33.6 &                                   0.816 &  46.5 &  37.5 &                                    0.886 &  48.7 &  36.4 &                         0.947 &  49.3 &  38.9 \\
             & $\ell_2$ & 78.0 &           0.913 &  44.2 &  35.4 &             0.903 &  48.0 &  39.4 &            0.956 &  48.9 &  40.9 &         0.970 &  49.8 &  40.8 \\
             & $\ell_2$+ Cosd (ours) & 78.0 &           \textbf{0.942} &  \textbf{45.2} &  \textbf{37.0} &             \textbf{0.934} &  \textbf{49.0} &  \textbf{41.1} &            \textbf{0.971} &  49.0 &  \textbf{42.5} &         \textbf{0.976} &  \textbf{50.9} &  \textbf{43.3} \\
\bottomrule
\end{tabular}
}
\end{table*}

\subsection{Experimental settings}
We used the CIFAR10 \cite{Krizhevsky_2009_17719} and ImageNet100 \cite{imagenet100,russakovsky2015imagenet} datasets to evaluate the robustness of our proposed regularization methods. The ImageNet100 dataset is a subset of the ImageNet-1k dataset with 100 of the 1K labels  selected. The train and test dataset contains 1.3K and 50 images for each class, respectively. We choose a three-layer custom convolutional neural network (LeNet) and ResNet18 \cite{he2016deep} for our experiments since our objective is not to achieve state-of-the-art accuracy for those datasets, but to evaluate the robustness of interpretation for popular models. We replaced all ReLU activations with Softplus$(\beta=3)$ as given by \cite{Dombrowski2019ExplanationsCB}, which makes the Hessian non-zero, as well as allows the model to have smoother geometry.
We set Adversarial Training on EXplanation (ATEX) \cite{tang2022defense}, triplet-based Input Gradient Alignment (IGA) \cite{singh2020attributional}, approximated norm of Hessian \cite{dombrowski2022towards}, $\ell_2$ robust criterion as our baselines. Our method is denoted by $\ell_2+$Cosd.  
We performed a grid search for selecting hyperparameters. For each hyperparameter, we trained the model three times from scratch and reported the mean value in Appendix C. Our code is available at \texttt{https://github.com/joshua840/RobustAGA}.

\subsubsection{Training details}
We applied two tricks to speed up our training. First, we sampled $\bmdx$ only once per iteration. Second, before calculating $\Gamma_{\nabx g_y}^{\ell_2}(\bfx, \bmdx)$ and $\Gamma_{\nabx g_y}^{\text{cos}}(\bfx, \bmdx)$ in (\ref{eq:crcr}), we treat $\nabx g_y(\bfx)$ as constant to prevent the gradient flows during back-propagation. The gradient of the regularization term is still propagated through $\nabx g_y(\bfx + \bmdx)$. We verified that the second technique accelerates the training speed by approximately 25\%, while there are no significant differences in the prediction accuracy and attribution robustness.

\subsection{Metrics}


\subsubsection{Random perturbation similarity}

We employ \textit{random perturbation similarity} (RPS) \cite{dombrowski2022towards}, which measures the similarity of the attribution at the given point $\bfx$ and the randomly perturbed point $\bfx+\bmdx$.
Formally, for given dataset $\mathcal{D}$, model $\bfg$, metric $\mathcal{S}$, attribution method $\mathcal{I}$, and noise level $\epsilon$, the RPS is defined as
\begin{align*}
    RP&S_{\bfg}( \mathcal{S},\mathcal{I}, \epsilon, \mathcal{D}) \\
    &= \frac{1}{N}\sum_{(\bfx,y) \in \mathcal{D}} \mathbb{E}_{\bmdx \sim \mathcal{U}_d ([-\epsilon, \epsilon]^d)} \mathcal{S}(\bfh_{g_y}^\mathcal{I}(\bfx+\bmdx), \bfh_{g_y}^\mathcal{I}(\bfx)),
\end{align*}
in which $N$ is the number of data tuples in $\mathcal{D}$ and $\mathcal{U}_d$ denotes the uniform distribution.
We performed Monte-Carlo sampling 10 times for each data tuple to approximate the expectation over $\bmdx$. We set $\epsilon$ as 4, 8, and 16. We employ cosine similarity, Pearson Correlation Coefficient (PCC), and Structural SIMilarity (SSIM) for the similarity metric $\mathcal{S}$, where PCC and SSIM are also used in previous literature \cite{dombrowski2022towards, adebayo2018sanity}. We will omit the $\bfg$ and $\mathcal{D}$ for simplicity.

\subsubsection{Insertion and Adv-Insertion game}

Another well-known strategy to measure the quality of attribution is to reconstruct the input pixels and observe the changes in output.
We employed the \textit{Insertion game} \cite{Petsiuk2018rise}, which observes the changes in categorical probability for the true label $y$ by inserting the input in the order of attribution score. In detail, we reconstruct the $\gamma$ ratio of input elements from the zero-valued image $\bfx_o$, in the order of the interpretation score $\bfh_{g_y}^\mathcal{I}(\bfx)$. To do this, we build a mask vector $\mathbf{m}_\gamma^\mathcal{I} \in \{0, 1\}^d$ such that each element satisfies $(\mathbf{m}_\gamma^\mathcal{I})_i = \mathbf{1}[(\bfh_{g_y}^\mathcal{I}(\bfx))_i > t_\gamma]$ and $\sum_i (\mathbf{m}_\gamma^\mathcal{I})_i / d = \gamma$ by carefully choosing the threshold $t_\gamma$. We denote the reconstructed input as $\bfx_\gamma^\mathcal{I}$ and define as $\bfx_\gamma^\mathcal{I} = \bfx \odot \mathbf{m}_\gamma^\mathcal{I} + \bfx_o \odot (\mathbf{1-m}_\gamma^\mathcal{I})$. Then, the average probability after insertion is defined as:
\begin{align*}
    \text{Insertion}_{\bfg}(\mathcal{I}, \mathcal{D}, \gamma) 
    &= \frac{1}{N}\sum_{(\bfx,y) \in \mathcal{D}} p_y(\bfx_\gamma^\mathcal{I}).
\end{align*}

We also propose \textit{Adv-Insertion} to measure the robustness of attributions against AAM. The overall process of Adv-Insertion is similar to normal Insertion, but Adv-Insertion determines the order of the input reconstruction by $\bfh_{g_y}^\mathcal{I}(\bfx_{adv})$, where the $\bfx_{adv}$ is derived from targeted AAM.
The A-Ins score will be high if the manipulated attribution $ \bfh_{g_y}^\mathcal{I}(\bfx_{adv})$ still well reflects the model behavior. 
For AAM in our experiments, we used PGD-$\ell_\infty(iter=100)$ where $\epsilon=2/255$ for ResNet18 and $\epsilon=4/255$ for LeNet. 
Also, we selected $\bfh_t$ as a frame image as shown in Figure \ref{fig:vis_random_paper} to minimize the overlap between the class object and target map.
We calculated Insertion and Adv-Insertion for each $\gamma\in\{0, 0.05, ..., 0.95, 1\}$ and posted the mean values over the reconstruction ratio $\gamma$.

\subsection{Quantitative and qualitative results}

Table \ref{tab:quantitative_sumary} shows quantitative results for four attribution methods and three robustness metrics. 
We denoted Insertion and Adv-Insertion as Ins and A-Ins, respectively.
We selected hyperparameters that have the highest RPS on the Grad attribution while allowing 1\% of test accuracy drops from CE only (without attribution regularization).For RPS, we only posted cosine similarity with $\epsilon=16$ results. 
All of the other results including cosine similarity, PCC, and SSIM with $\epsilon=4, 8, 16$, and the other hyperparameters are given in Appendix C.
Note that we did not evaluate the results for IGA and ATEX in ResNet18 since they have high computational costs. 



\begin{figure}[ht]
     \centering
     \begin{subfigure}[b]{\columnwidth}
         \centering
         \includegraphics[width=\textwidth]{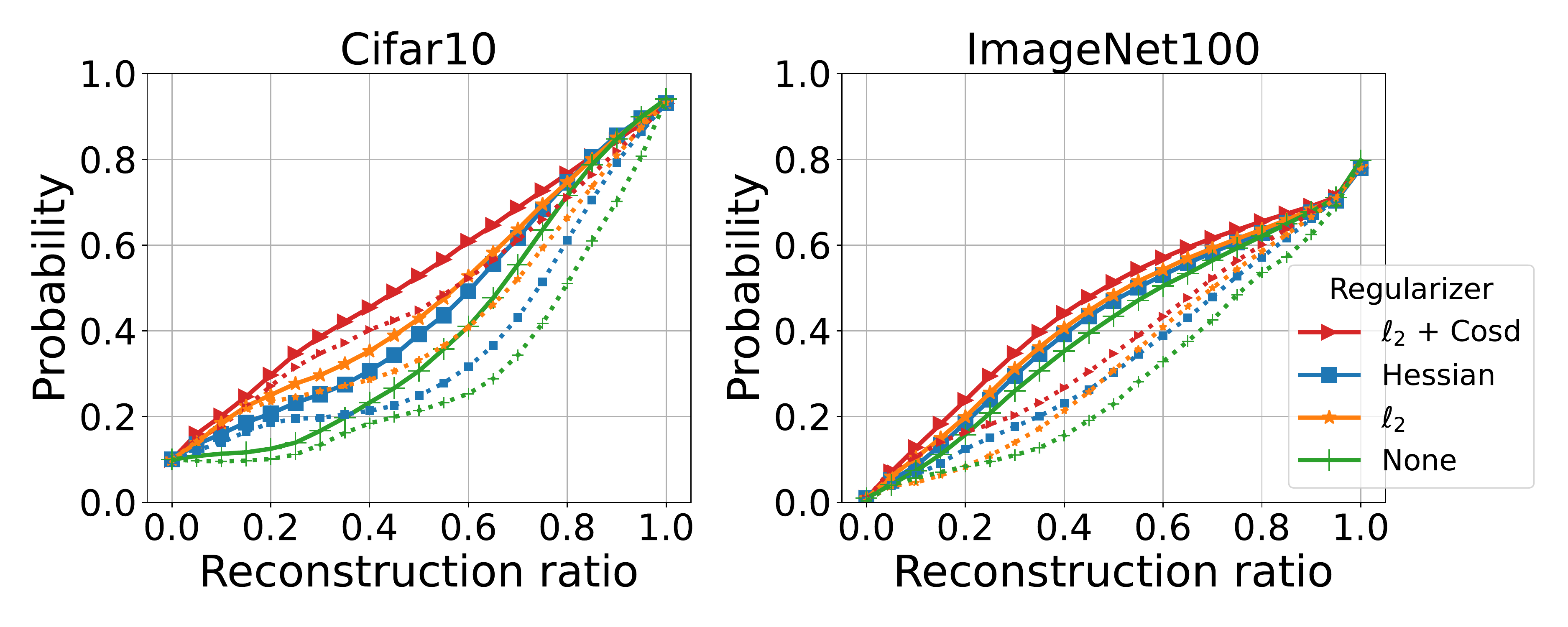}
         \caption{Insertion (solid line) and Adv-Insertion (dotted line) curves for CIFAR10 and ImageNet. ResNet18 and Grad explanation are used for both graphs.}
         \label{fig:insertion1}
     \end{subfigure}
    
     \begin{subfigure}[b]{\columnwidth}
         \centering
         \includegraphics[width=\textwidth]{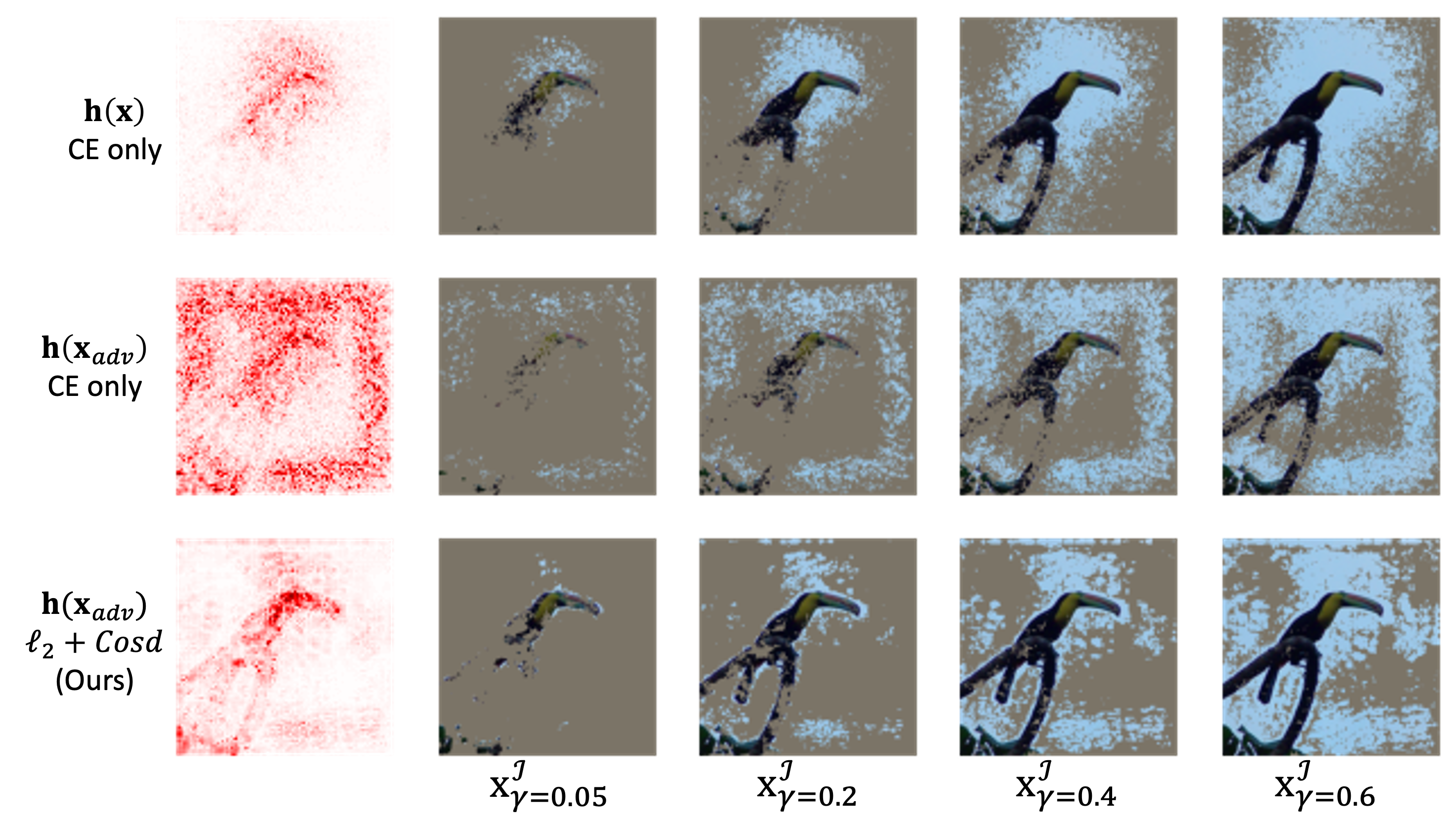}
         \caption{Examples of reconstructed images that are used in the Insertion and Adv-Insertion metrics. 
         \textbf{Column 1}: Visualizations of Grad explanation, which determines the reconstruction order.
         \textbf{Column 2 $\sim$ 5}: Reconstructed images where $\gamma=\{0.05,0.2,0.4,0.6\}$.
         }
         \label{fig:insertion2}
     \end{subfigure}
     
    \caption{Insertion and Adv-Insertion metrics.}
    \label{fig:insertion}
\end{figure}

\subsubsection{Robustness on the random perturbation}
Overall, our method shows the highest RPS on CIFAR10 and ImageNet100 for almost all attribution methods, which underscores that the model trained by our method has smoother geometry on average than the other methods.
In the case of ATEX, the self-knowledge distillation gives high accuracy gain but the robustness of the attribution is poor. 
Interestingly, our method has better robustness results than IGA, which is computationally expensive with inner maximization computations.
Also, it turns out that ATEX, IGA, and Hessian are worse than not only our method but also the method with only $\ell_2$, which needs much less computation time and memory.



\subsubsection{Insertion and Adv-Insertion}
Our Insertion and Adv-Insertion game results are given in both Table \ref{tab:quantitative_sumary} and Figure \ref{fig:insertion}.
In Table \ref{tab:quantitative_sumary}, the model trained by our method achieves the best or comparable Insertion and Adv-Insertion scores. The tendency of Adv-Insertion is similar for different values of the $\epsilon$, which are given in Appendix C.  Especially, the \textit{Adv-Insertion} score of the model trained on CIFAR10-ResNet18 with our method is comparable to the \textit{Insertion} score of the model trained with $\ell_2$ regularization, which highlights the robustness of our method against the adversarial input.
In Figure \ref{fig:insertion}(a), we can observe that our $\ell_2$+Cosd method has the highest probability all over the reconstruction ratio. Qualitatively, Figure \ref{fig:insertion}(b) shows that the main object part is well preserved in the reconstructed image for our method since $\bfh(\bfx_{adv})$ still mainly highlights the main object even for the adversarial input, in contrast to the CE only training. 



\begin{figure}[ht]
     \centering
     \begin{subfigure}[t]{0.64\columnwidth}
         \centering
         \includegraphics[width=\textwidth]{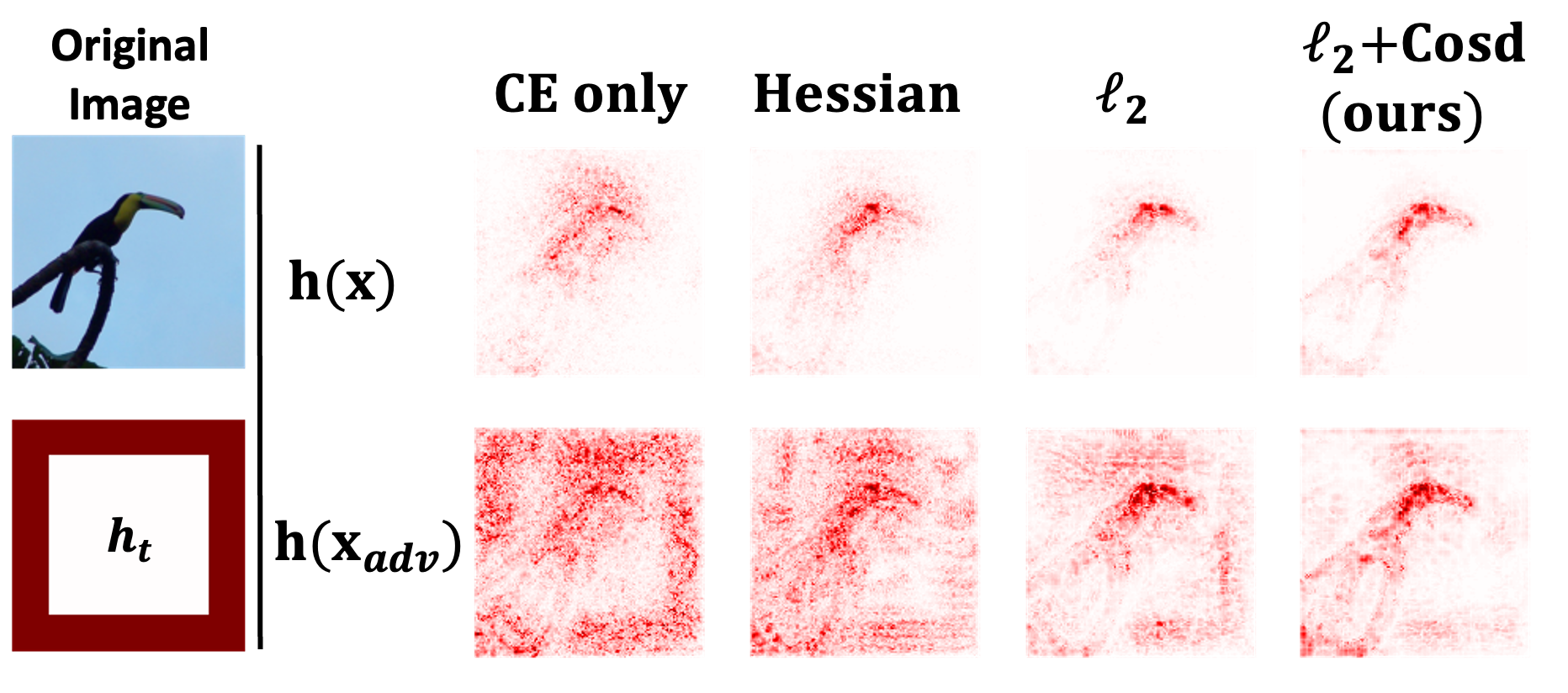}
         \caption{Attribution Visualization}
         \label{fig:vis_random_paper}
     \end{subfigure}
      \begin{subfigure}[t]{0.35\columnwidth}
         \centering
         \includegraphics[width=\textwidth]{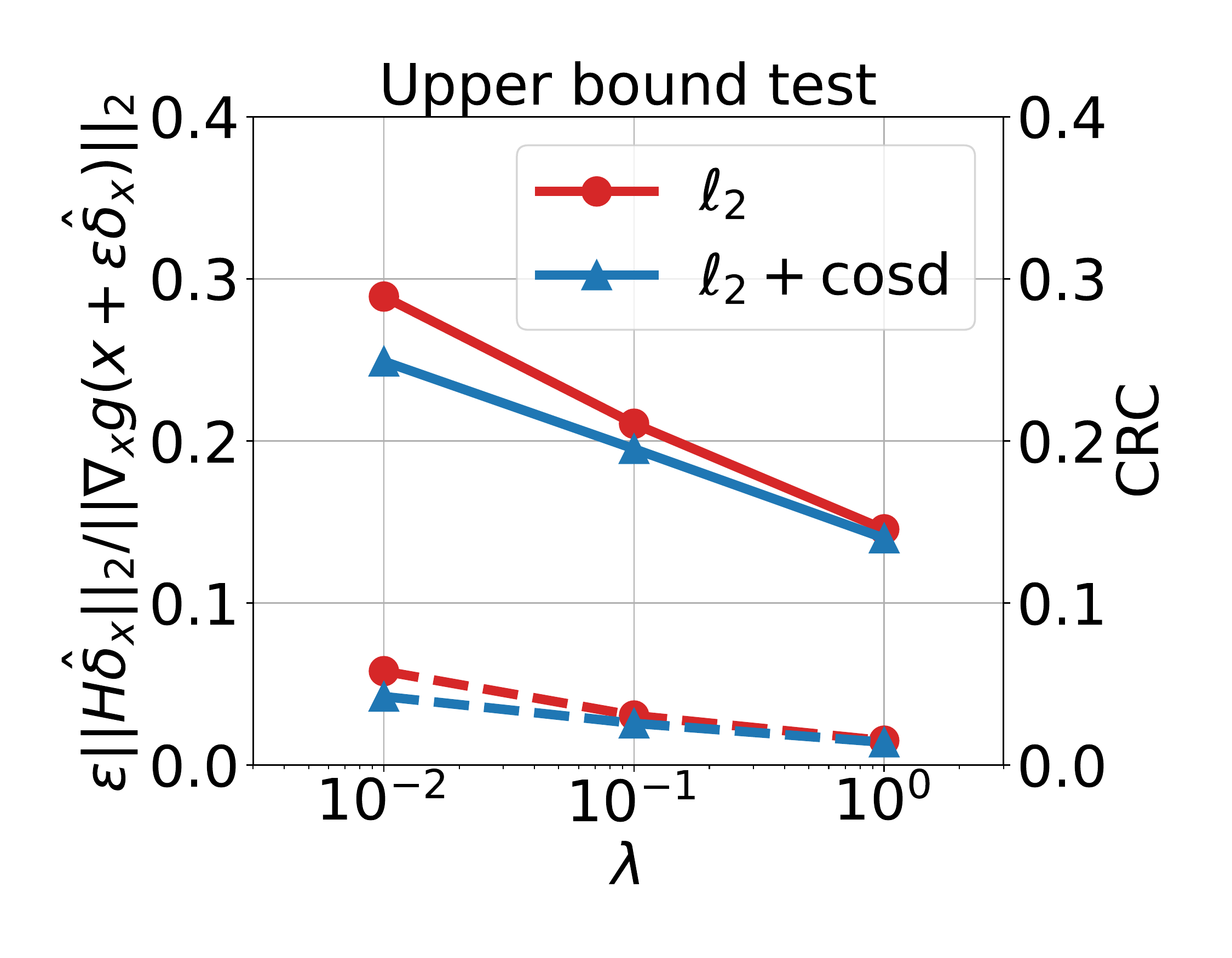}
         \caption{ Upper bound test}
         \label{fig:upperbound}
     \end{subfigure}
     
    \caption{ (a) \textbf{Row 1}: Visualization of attribution maps for each of the four different models, CE only, Hessian, $\ell_2$ and $\ell_2$ + Cosd. \textbf{Row 2}: Adversarially attacked images, \textit{i.e., }$\bfx_{adv}$ with targeted AAM with $\bfh_t$, are made and the attacked attribution maps are generated for each model. (b) \textbf{Solid}: Upper bound in (\ref{eq:upper bound}) without the $\mathcal{O}(\epsilon^2)$ term, \textbf{Dotted}: CRC in Def.\ref{def:crc}.}
    \label{fig:3}
\end{figure}




\subsubsection{Qualitative results}
Figure \ref{fig:vis_random_paper} shows saliency map visualization. We visualize the Grad attributions for both the source image $\bfx$ and the manipulated images $\bfx_{adv}$, where $\bfx_{adv}$ is derived from targeted AAM with PGD-${\ell_\infty}(\epsilon=2/255, iter=100)$ applied. The attribution of the manipulated image with our proposed method shows a much more robust appearance than those with the other methods using only Hessian or $\ell_2$; \textit{i.e.}, ours shows attribution that is clearly sparse and highlights the main object, while others include the spurious target attribution $h_t$.

\begin{table}[ht]
\centering
\caption{Training complexity}
\label{tab:complexity}
\resizebox{\columnwidth}{!}{
\begin{tabular}{l|ll|ll|ll}
\toprule
          & \multicolumn{2}{c|}{\shortstack{CIFAR10\\LeNet}}    & \multicolumn{2}{c|}{\shortstack{CIFAR10\\ResNet18}}                      & \multicolumn{2}{c}{\shortstack{ImageNet100\\ResNet18}}                \\
\midrule
Regularizer        &    Time & Memory  &    Time & Memory  &    Time & Memory \\
\midrule
CE only                               & $\times$ 1.0         & $\times$ 1.0  & $\times$ 1.0         & $\times$ 1.0   & $\times$ 1.0         & $\times$ 1.0   \\
Hessian     & $\times$ 4.25    & $\times$ 4.36   & $\times$ 8.95     & $\times$ 6.62   & $\times$ 13.88       & $\times$ 4.06  \\
 ATEX $(3^{rd}$ step)  & $\times$ 2.25    &  $\times$ 2.81 &  $\times$ 2.22    & $\times$ 3.16 &  -    &  -  \\
IGA  & $\times$  31.0    & $\times$ 6.55     & $\times$  26.45  & $\times$  6.74   &  -    &  -   \\
$\ell_2$+Cosd  (ours)       &         $\times$ 2.75     & $\times$ 2.64      & $\times$ 4.10         & $\times$ 2.65   & $\times$ 4.94        & $\times$ 2.19  \\

\bottomrule
\end{tabular}

}
\end{table}

\section{Discussion}
\subsubsection{Upper bounds} In Figure \ref{fig:upperbound}, we plot $\frac{\epsilon ||\hesg(\bfx)\hbmdx||_2 }{||\nabx g(\bfx+\epsilon\hbmdx)||_2}$, a slightly tighter term than what is given as an upper bound in Equation \ref{eq:upper bound}, and the cosine distance for models that trained with $\ell_2$ and $\ell_2$+Cosd on the CIFAR10-LeNet. The figure shows that the upper bound on CRC is well aligned with the cosine distance. 


\subsubsection{Training speed}
Table \ref{tab:complexity} shows the memory and time requirements of the regularization methods used in Table \ref{tab:quantitative_sumary}. We set the CE-only model as $\times$1 and reported the relative time and memory consumption. Our method is much more efficient than the other regularization methods, from both time and memory perspectives.
Note that ATEX has three steps, but we only measured the time and accuracy for the third step.

\subsubsection{Second order derivatives}
Hessian regularizer is only applicable when the model is twice differentiable. Hence, the Hessian calculation cannot be performed on a model with ReLU activations, because $ \nabla_\bfx^2ReLU(\bfw^T\bfx) = \mathbf{0}$. 
However, the ReLU network can be trained with the $\ell_2$ or cosine regularization method, because $\nabla_\bfw \nabla_\bfx ReLU(\bfw^T\bfx)\neq \mathbf{0}$. 

\subsubsection{Robustness of SmoothGrad}
We checked that SmoothGrad \cite{smilkov2017smoothgrad} is robust to AAM, because the ball of sampling random perturbation of SmoothGrad is much larger than the ball of adversarial perturbation. This is also mentioned in \cite{Dombrowski2019ExplanationsCB}.

\subsubsection{Instability of CRC regularization} We observed that the attribution map of a model trained with CRC regularization with a large coefficient can be broken. Namely, the attribution map always highlights the edges regardless of input. In this case, the RPS is very high, but the Insertion or Adv-Insertion value is low. The visualizations of the broken attributions are given in Appendix C.

\section{Concluding Remarks}


We promote the alignment of local gradients by suggesting a normalization invariant criterion. Our new attribution robust criterion overcomes previous limitations and our combined regularization method achieves better robustness and explanation quality in large-scale settings with lower computation costs than previous methods.

Even though our proposed method achieved better results and faster training speed than the baselines, there exist some limitations. 
First, our proposed regularization method is slower than the ordinary training with only cross-entropy loss. The requirement for tuning several hyperparameters is another limitation. 

\section{Acknowledgement}
This work was supported in part by the New Faculty Startup Fund from Seoul National University, NRF grants [NRF-2021M3E5D2A01024795] and IITP grants [RS-2022-00155958, No.2021-0-01343, No.2021-0-02068, No.2022-0-00959] funded by the Korean government. AW acknowledges support from a Turing AI Fellowship under EPSRC grant EP/V025279/1, The Alan Turing Institute, and the Leverhulme Trust via CFI.


\bibliography{custom, bibfile}



\end{document}


\maketitle

\section{A. Training Details} 

\subsection{Hyperparameters of training LeNet with CIFAR10 dataset} \label{app:A1}
\begin{itemize}
    \item \textbf{Regularizer}: [$\ell_2$ + Cosd , $\ell_2$, Hessian, ATEX, IGA]
    \item \textbf{Model}: LeNet
    \item \textbf{Optimizer}: IGA: SGD(learning-rate=0.1, weight-decay=2e-4)  Others: Adam(learning-rate=1e-3, weight-decay=4e-5), 
    \item \textbf{Training batch size}: 128
    \item \textbf{Epochs}: 200
    \item \textbf{Learning rate decay}: IGA: 1/10 at [50,80,150] epochs  Others:1/10 at [100, 150] epochs
    \item \textbf{Activation}:  IGA: Softplus$(\beta=50)$  Others: Softplus$(\beta=3)$, 
    \item \textbf{Regularization constants}: 
    \begin{itemize}
        \item $\ell_2$ + cosd: $\lambda \times \lambda_c$ = \{0, 1e-2, \textbf{ 1e-1}, 1\} $\times$ \{1e-2, 1e-1, \textbf{1}\}
        \item $\ell_2$ : $\lambda$ = \{1e-2, \textbf{1e-1}, 1\}
        \item Hessian: $\lambda$ = \{1e-6, 1e-5, 1e-4, \textbf{1e-3}, 1e-2, 1e-1, 1\}
        \item ATEX: $\lambda \times \epsilon$ = \{0.1, 0.3, 1, \textbf{3}\} $\times$ \{\textbf{2}, 4\}
        \item IGA: $\lambda$ = \{1e-1, 5e-1, \textbf{1}\}
    \end{itemize}
    
\end{itemize}

\subsection{Hyperparameters of training ResNet with CIFAR10 dataset}
\begin{itemize}
    \item \textbf{Regularizer}: [$\ell_2$ + Cosd , $\ell_2$, Hessian]
    \item \textbf{Model}: ResNet18
    \item \textbf{Optimizer}: AdamW(learning-rate=1e-3, weight-decay=1e-2)
    \item \textbf{Training batch size}: 128
    \item \textbf{Epochs}: 200
    \item \textbf{Learning rate decay}: 1/10 at [100, 150] epochs
    \item \textbf{Activation}:  Softplus$(\beta=3)$
    \item \textbf{Regularization constants}: 
    \begin{itemize}
        \item $\ell_2$ + cosd: $\lambda \times \lambda_c$ = \{0, 1e-2, \textbf{1e-1}, 1\} $\times$ \{1e-2, 1e-1, \textbf{1}\}
        \item $\ell_2$ : $\lambda$ = \{1e-2, 3e-2, 1e-1, 3e-1, \textbf{1}, 3\}
        \item Hessian: $\lambda$ = \{1e-6, \textbf{1e-5}, 1e-4, 1e-3\}
    \end{itemize}
    
\end{itemize}

\subsection{Hyperparameters of training ResNet with ImageNet100 dataset}
\begin{itemize}
    \item \textbf{Regularizer}: [$\ell_2$ + Cosd , $\ell_2$, Hessian]
    \item \textbf{Model}: ResNet18
    \item \textbf{Optimizer}: AdamW(learning-rate=1e-3, weight-decay=1e-2)
    \item \textbf{Training batch size}: 128
    \item \textbf{Epochs}: 90
    \item \textbf{Learning rate decay}: 1/10 at [40, 60, 80] epochs
    \item \textbf{Activation}: Softplus$(\beta=3)$
    \item \textbf{Regularization constants}: 
    \begin{itemize}
        \item $\ell_2$ + cosd: $\lambda \times \lambda_c$ = \{0, 1e-2, 1e-1, \textbf{1}\} $\times$ \{1e-2, 1e-1, \textbf{1}\}
        \item $\ell_2$ : $\lambda$ = \{1e-2, 3e-2, 1e-1, \textbf{3e-1}, 1\}
        \item Hessian: $\lambda$ = \{1e-5, \textbf{1e-4}, 1e-3\}
    \end{itemize}
\end{itemize}

* \textbf{Bold}: Selected hyperparameters in the main paper.

\subsection{Architecture}

For our LeNet on CIFAR10, we used a model consisting of four convolution layers, two max-pooling layers, and two fully connected layers. This architecture is exactly the same as given in \cite{dombrowski2022towards}
\begin{verbatim}
LeNet(
    Conv2d(3, 32, kernel_size=(3, 3), stride=(1, 1), padding=(1, 1))
    ActivationFunction()
    Conv2d(32, 32, kernel_size=(3, 3), stride=(1, 1), padding=(1, 1))
    ActivationFunction()
    MaxPool2d(kernel_size=2, stride=2, padding=0, dilation=1)
    Conv2d(32, 64, kernel_size=(3, 3), stride=(1, 1), padding=(1, 1))
    ActivationFunction()
    Conv2d(64, 64, kernel_size=(3, 3), stride=(1, 1), padding=(1, 1))
    ActivationFunction()
    MaxPool2d(kernel_size=2, stride=2, padding=0, dilation=1)
    Flatten()
    Linear(in_features=4096, out_features=256, bias=True)
    ActivationFunction()
    Linear(in_features=256, out_features=10, bias=True)
)
\end{verbatim}

For ResNet18 on CIFAR10, we adopted the widely used open-source implementation, which is given in (\texttt{https://github.com/kuangliu/pytorch-cifar/blob/master/models/resnet.py}.)
For ResNet18 on ImageNet, we used a model in a Torchvision framework.

\subsection{Experiment setting in details}
We used the CIFAR10  \cite{Krizhevsky_2009_17719} and ImageNet100 \cite{imagenet100,russakovsky2015imagenet} datasets to train and measure the robustness of our proposed methods. The ImageNet100 dataset is a subset of the ImageNet-1k dataset with 100 of the 1K labels  selected. The train and test dataset contains 1.3K and 50 images for each class, respectively. For the CIFAR10 dataset, we employed a three-layer custom convolutional neural network (LeNet) and ResNet18 \cite{he2016deep} models and trained them for 200 epochs from scratch. We decayed the learning rates by multiplying 0.1 at each 100 and 150 epochs. 
For the ImageNet100\cite{imagenet100}, 
we trained ResNet18 for 90 epochs from scratch with learning rate decay at 40, 60, and 80 epochs. 

For ResNet18 models, we used the AdamW optimizer \cite{AdamW}, where learning rate, weight decays, and batch-sizes are 1e-3, 1e-2, and 128, respectively. For LeNet models, we used the Adam optimizer \cite{kingma2014adam}, where learning rate, weight decays, and batch-sizes are 1e-3, 4e-5, and 128, respectively. 
We performed a grid search on the hyperparameters to train the models, using Softplus$(\beta=3)$ for the activation functions.  A combination of both cosine distance and  $\ell_2$ (\textit{i.e.,} our proposed method), $\ell_2$, and Hessian norm approximation were used as regularizers. Additionally, we used ATEX and IGA methods in CIFAR10, LeNet. We used different lists of $\lambda$ for each regularizer and $\lambda_c$ for methods combined with both cosine distance and  $\ell_2$ as given in Appendix.
We trained each model using NVIDIA A100 and RTX A5000 GPUs with the PyTorch framework.

We carefully implemented the attribution methods used in this paper to utilize the back-propagate from the feature attributions. 
In case of Grad, xGrad, and GBP, we checked that our implementation returns exactly the same attribution as in the Captum \cite{kokhlikyan2020captum} which is a well-known framework but does not offer the back-propagation through attributions. 
In the case of LRP, we followed the Zennit \cite{anders2021software} implement  and also checked the same output for VGG models. Then, we applied the LRP rules for ResNet18 model.

\subsection{Regularization methods}

\subsubsection{Adversarial training on explanations (ATEX)}
ATEX is a distillation-based training method for robust explanations that do not require a second-order calculation of gradient. They choose which inputs to use for distillation by considering the geometry of the pre-trained model. By assuming that a model $f$ is trained with ordinary cross-entropy loss on the training dataset, they first define a set of inputs that is parallel to the direction of attribution, which is denoted and defined as
$$\mathcal{I}(\bfx) = \{\bfx^i | \bfx^i = \bfx + \Delta \bfh_f(\bfx) / ||\bfh_f(\bfx)||_2, \epsilon \leq \Delta \leq \epsilon\},$$
where $\epsilon$ is a hyperparameter. We set the $\bfh_f$ as SmoothGrad \cite{smilkov2017smoothgrad} as given in \cite{tang2022defense}.
Also, they define another set of perturbations $\mathcal{P}(\bfx^i)$, which is defined as
$$\mathcal{P}(\bfx^i) = \{\bfx^p | \bfx^p = \bfx^i + \Delta \bfh_f^\perp(\bfx) / ||\bfh_f^\perp(\bfx)||_2, \epsilon \leq \Delta \leq \epsilon\},$$
where $\bfh_f^\perp(\cdot)$ denotes a perpendicular direction from $\bfh_f$.
Then, a loss function of ATEX is denoted and defined as
$$\mathcal{L}((\bfx, y); \Theta_f, \Theta_g) = KL(\bfg(\bfx)||\bff(\bfx)) + \lambda \sum_{\bfx^i\sim\mathcal{I}(\bfx)} \sum_{\bfx^p\sim\mathcal{P}(\bfx^i)} KL(\bfg(\bfx^p)||\bff(\bfx^i)),$$
where $\Theta_f$ is a parameter of model $\bff$, $\Theta_g$ is a parameter of model $\bfg$ to be updated. We trained models with $(\lambda \times \epsilon) \in \{0.1, 0.3, 1, 3\} \times \{2, 4\}$. 
Note that this method is extremely inefficient if there are data augmentation during training because the $\bfh_f(\bfx)$ should be calculated for every training iteration. Hence we only evaluated this method for CIFAR10-LeNet.

\subsubsection{Input-gradient spatial alignment (IGA)}
IGA is a training method that aims to minimize the upper bound of spatial correlation between the input and attribution map. This method optimizes the $\mathcal{L}_{attr}$ which is a triplet loss with a soft margin on cosine distance between $\bfg^{i^*}$ and $\bfx$. It is known that a model trained with this method is robust not only in adversarial model manipulation but also in adversarial perturbations. Their soft-margin triplet loss function is denoted and defined as
$$\mathcal{L}_{attr}(\bfx,y)=log(1+exp(-(\bfd(\nabx\bfg_{i^*}(\bfx),\bfx),\bfd(\nabx\bfg_{y}(\bfx),\bfx) ))),$$
where $\bfd(\nabx\bfg_{i^*}(\bfx),\bfx)$ is cosine distance,$i^* = \argmax_{y \neq i}\bff_i(\bfx)$.
Then, the overall loss function is defined as
$$\mathcal{L}(\bfx,y)= \mathcal{L}_{ce}(\bfx+\bmdx, y) +  \lambda\mathcal{L}_{attr}(\bfx+\bmdx, y),$$ where $\bmdx = \argmax_{||\bmdx||_\infty < \epsilon} \mathcal{L}_{attr}(\bfx+\bmdx,y)$.
We trained models with $\epsilon\in \{0.4, 1\}$ to find models where the accuracy is not lower than 1\% from the CE-only model. Note that this method needs many computation resources in large-scale datasets. Hence, we only use CIFAR10-LeNet.

\section{B. Proofs}
\subsection{Proof of Theorem 3.6}
\label{app:proof_upper_bound}
The proof of Theorem 3.6 is given by
\begin{align}
    2&\Gamma^c_{\nabla_g}(\bfx, \epsilon\hbmdx) \nonumber\\
    &= 1-\text{cossim}\left(\nabx g(\bfx+\epsilon\hbmdx), \nabx g(\bfx))\right) \nonumber\\
    &=1-\frac{<\nabx g(\bfx+\epsilon\hbmdx), \nabx g(\bfx)>} 
    {||\nabx g(\bfx+\epsilon\hbmdx)||_2 \cdot ||\nabx g(\bfx)||_2} \nonumber\\
    &=1-\frac{<\nabx g(\bfx+\epsilon\hbmdx), \mathbf{e}_{\nabx g(\bfx)}>} 
    {||\nabx g(\bfx+\epsilon\hbmdx)||_2} \nonumber\\
    &=\frac{||\nabx g(\bfx+\epsilon\hbmdx)||_2 - <\nabx g(\bfx+\epsilon\hbmdx), \mathbf{e}_{\nabx g(\bfx)}>}
    {||\nabx g(\bfx+\epsilon\hbmdx)||_2} \nonumber\\
    &=\frac{||\nabx g(\bfx)+\epsilon\hesg\nonumber(\bfx)\hbmdx+\mathcal{O}(\epsilon^2)\mathbf{v}||_2 - <\nabx g(\bfx)+\epsilon\hesg(\bfx)\hbmdx+\mathcal{O}(\epsilon^2)\mathbf{v}, \mathbf{e}_{\nabx g(\bfx)}>}{||\nabx g(\bfx+\epsilon\hbmdx)||_2}\\
    &\leq \frac{||\nabx g(\bfx)||_2 + ||\epsilon\hesg(\bfx)\hbmdx||_2 + \mathcal{O}(\epsilon^2)||\mathbf{v}||_2 - ||\nabx g(\bfx)||_2 - <\epsilon\hesg(\bfx)\hbmdx, \mathbf{e}_{\nabx g(\bfx)}> - <\mathcal{O}(\epsilon^2)\mathbf{v}, \mathbf{e}_{\nabx g(\bfx)}>}
    {||\nabx g(\bfx+\epsilon\hbmdx)||_2} \label{eq2}\\
    &=\frac{||\epsilon\hesg(\bfx)\hbmdx||_2 - <\epsilon\hesg(\bfx)\hbmdx, \mathbf{e}_{\nabx g(\bfx)}>  + \mathcal{O}(\epsilon^2) }
    {||\nabx g(\bfx+\epsilon\hbmdx)||_2\nonumber}\\
    &=\frac{||\epsilon\hesg(\bfx)\hbmdx||_2 - ||\epsilon\hesg(\bfx)\hbmdx||_2<\mathbf{e}_{\hesg(\bfx)\hbmdx}, \mathbf{e}_{\nabx g(\bfx)}> + \mathcal{O}(\epsilon^2) }
    {||\nabx g(\bfx+\epsilon\hbmdx)||_2} \nonumber\\
    &= \frac{||\epsilon\hesg(\bfx)\hbmdx||_2 <\mathbf{e}_{\hesg(\bfx)\hbmdx}, \mathbf{e}_{\hesg(\bfx)\hbmdx} - \mathbf{e}_{\nabx g(\bfx)}> + \mathcal{O}(\epsilon^2) }
    {||\nabx g(\bfx+\epsilon\hbmdx)||_2} \nonumber\\
    &\leq 2\frac{\epsilon||\hesg(\bfx) \hbmdx||_2 + \mathcal{O}(\epsilon^2) }
    {||\nabx g(\bfx+\epsilon\hbmdx)||_2}, \label{eq3}\\
    &\leq 2\frac{\epsilon||\hesg(\bfx)||_2 + \mathcal{O}(\epsilon^2) }
    {||\nabx g(\bfx+\epsilon\hbmdx)||_2},\label{eq4}
\end{align}
in which $\hbmdx$ is an unit vector, $\mathbf{e}_{\hesg(\bfx)\hbmdx}$ and $\mathbf{e}_{\nabx g(\bfx)}$ are unit vectors,  $\hesg(\bfx)\hbmdx/||\hesg(\bfx)\hbmdx||_2$ and $\nabx g(\bfx) /||\nabx g(\bfx)||_2$, respectively. For the proof, we used the Taylor expansion
$$
\nabx g(\bfx + \epsilon\hbmdx) = \nabx g(\bfx) + \hesg(\bfx)\epsilon\hbmdx + \mathcal{O}(\epsilon^2)\mathbf{v}
$$
for some vector $\mathbf{v}$, triangle inequality for (\ref{eq2}), and the fact 
\begin{align*}
       &<\mathbf{e}_{\hesg(\bfx)\hbmdx}, \mathbf{e}_{\hesg(\bfx)\hbmdx} - \mathbf{e}_{\nabx g(\bfx)}> \leq 2\\
    &||\mathbf{A}\bfx||_2 \leq ||\mathbf{A}||_2||\bfx||_2 \\
\end{align*}
for (\ref{eq3}) and (\ref{eq4}), respectively. \qed

\subsection{Proof of the non-negative homogeneity of feed-forward ReLU network}
\label{app:proof_relu_nnh}
Without loss of generality, we set $\bmth^{(\ell)} = \bfW^{(\ell)}$, where each $\bfW^{(\ell)}$ represents a weight matrix of the $\ell-$th fully connected layer. We represent a FNN with ReLU activation as
\begin{align*}
    \bff(\bfx;\Theta) = \bfW^{(L)} (\bfW^{(L-1)} ... (\bfW^{(2)}(\bfW^{(1)}\bfx)_+)_+ ...)_+,
\end{align*}
where $(\bfx)_+ = max(\bfx, 0)$ is a ReLU function. We include the bias term $\mathbf{b}$ of the fully connected layer in its weight matrix by re-writing $\bfx \coloneqq [\bfx^T, 1]^T$ and $\bfW \coloneqq [\bfW, \mathbf{b}]$.
Then, from $max(\alpha\bfx, 0) = \alpha max(\bfx, 0)$ and simple factorization, we can show that the following equation holds.
\begin{align*}
    \bff(\bfx;\{\alpha\bmth_1, \bmth_2, \bmth_3, ..., \bmth_L\}) &= \bfW^{(L)} (\bfW^{(L-1)} ... (\bfW^{(2)}(\alpha\bfW^{(1)}\bfx)_+)_+ ...)_+ \\
    &= \bfW^{(L)} (\bfW^{(L-1)} ... (\alpha\bfW^{(2)}(\bfW^{(1)}\bfx)_+)_+ ...)_+ \\
    & ...
    \\
    &= \alpha\bfW^{(L)} (\bfW^{(L-1)} ... (\bfW^{(2)}(\bfW^{(1)}\bfx)_+)_+ ...)_+ \\
    &= \alpha\bff(\bfx;\Theta)
\end{align*}

In the case of Softplus activation function, we cannot use the factorization, because of Softplus$(\alpha\bfx) \ne \alpha$ Softplus$(\bfx)$.

\section{C. Full experimental results} 

\subsection{Quantitative results}
All of the quantitative results with every hyperparameter for each dataset, model, and regularization method are given in Table \ref{tab:appendix_rps1}, \ref{tab:appendix_rps2}, and \ref{tab:appendix_insertion}. Table \ref{tab:appendix_rps1} and \ref{tab:appendix_rps2} represents the RPS for each model trained with different regularization methods and hyperparameters, where $\epsilon \in \{4/255, 8/255, 16/255\}$. Table \ref{tab:appendix_insertion} represents  Insertion and Adv-Insertion score of Grad, xGrad, GBP, and LRP for every regularization method and hyperparameters. 
Note that LRP on the model trained on Hessian regularization with $\lambda = 1e-3$ is always zero. It is because the model has only minus weights at the last fully-connected layer, and the $z+$ rule only calculates using the plus weights.


\subsection{Qualitative results}

\begin{figure}[ht]
\vskip 0.2in
\begin{center}
\centerline{\includegraphics[width=\columnwidth]{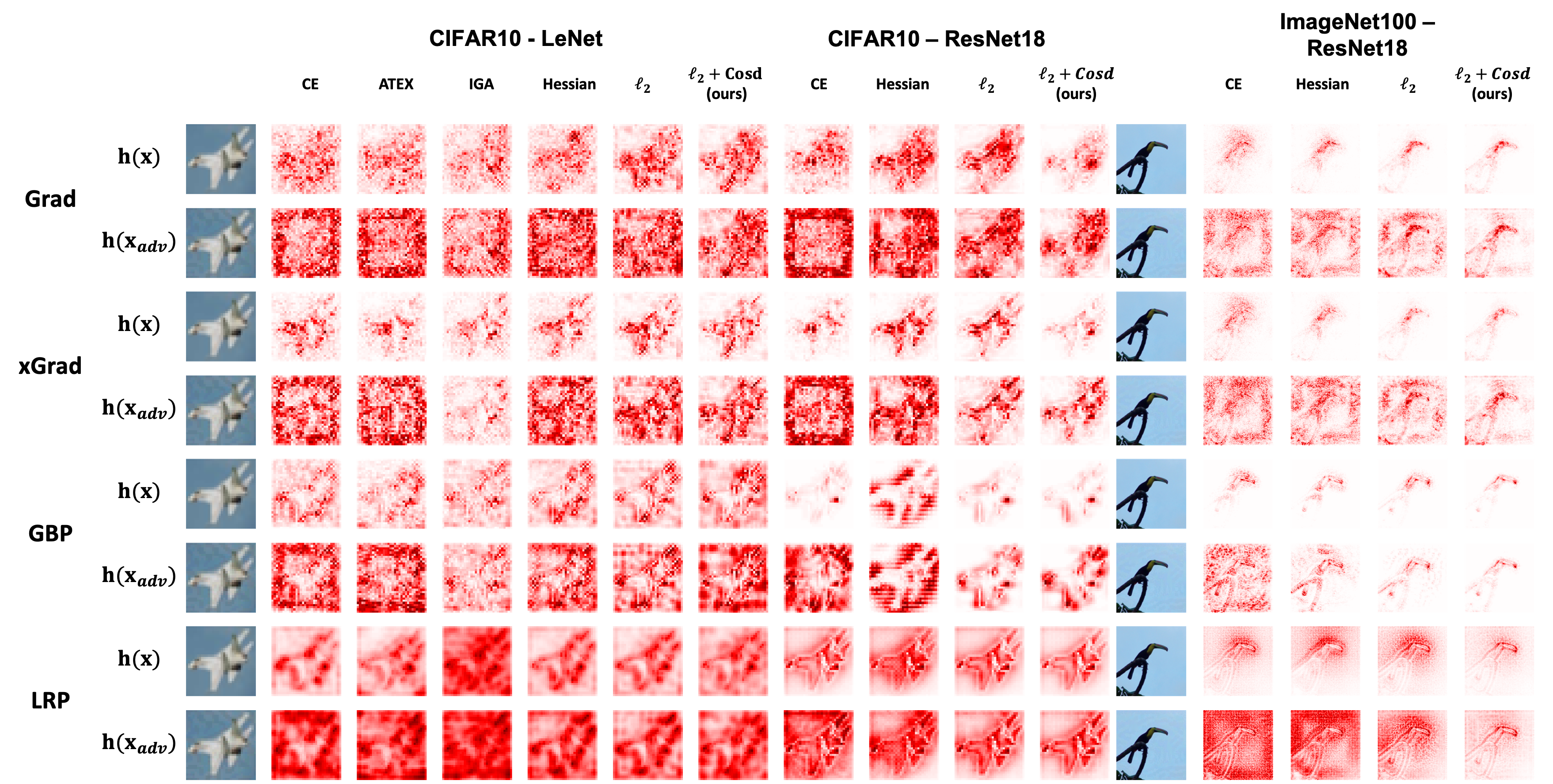}}
\caption{Visualization of selected models' attributions for original and adversarially manipulated images with all attribution methods. The AAM has been performed by PGD$_{\ell_\infty}(\epsilon=4)$ on CIFAR10 and PGD$_{\ell_\infty}(\epsilon=2)$ on ImageNet100 }
\label{fig:vis_all}
\end{center}
\vskip -0.2in
\end{figure}

\begin{figure}[ht]
\vskip 0.2in
\begin{center}
\centerline{\includegraphics[width=\columnwidth]{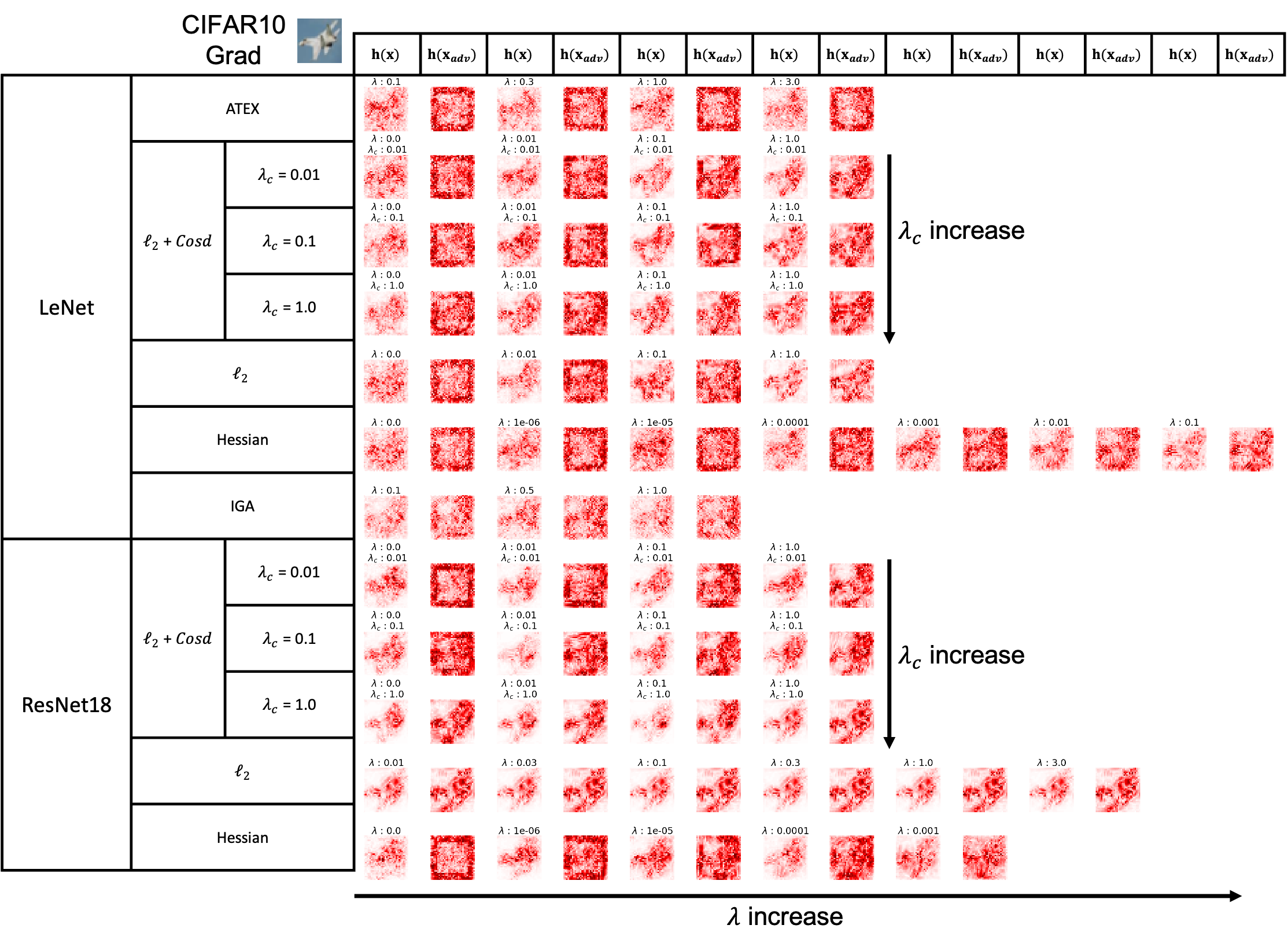}}
\caption{Visualization of the original and manipulated Gradient attribution for a CIFAR10 image with each model. The CE model corresponds to the $\ell_2$ and Hessian with $\lambda=0$. The AAM has been performed by PGD$_{\ell_\infty}(\epsilon=4)$}
\label{fig:cifar10_img1}
\end{center}
\vskip -0.2in
\end{figure}

\begin{figure}[ht]
\vskip 0.2in
\begin{center}
\centerline{\includegraphics[width=\columnwidth]{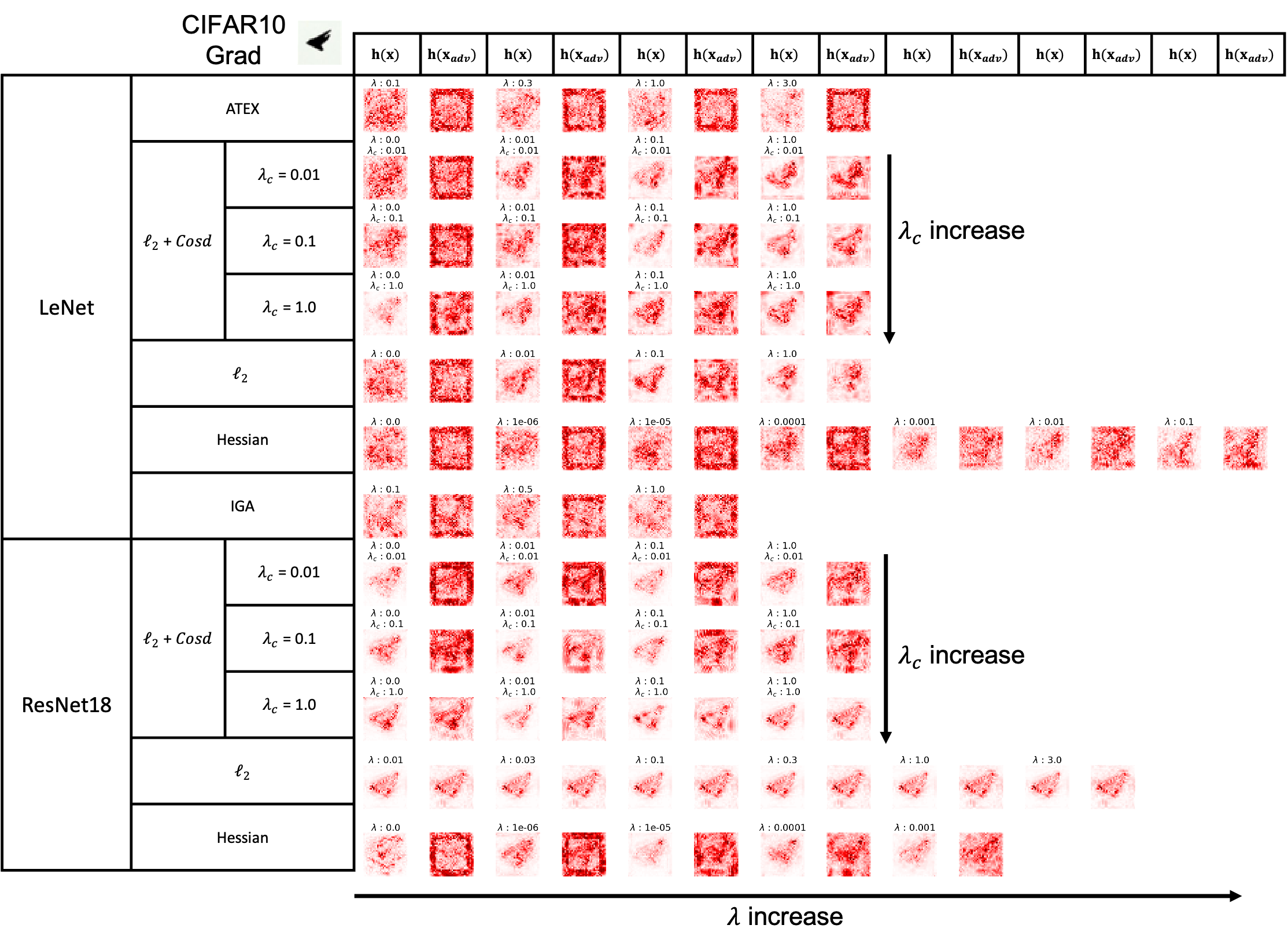}}
\caption{Visualization of the original and manipulated Gradient attribution for a CIFAR10 image with each model. The CE model corresponds to the $\ell_2$ and Hessian with $\lambda=0$. The AAM has been performed by PGD$_{\ell_\infty}(\epsilon=4)$}
\label{fig:cifar10_img2}
\end{center}
\vskip -0.2in
\end{figure}

\begin{figure}[ht]
\vskip 0.2in
\begin{center}
\centerline{\includegraphics[width=\columnwidth]{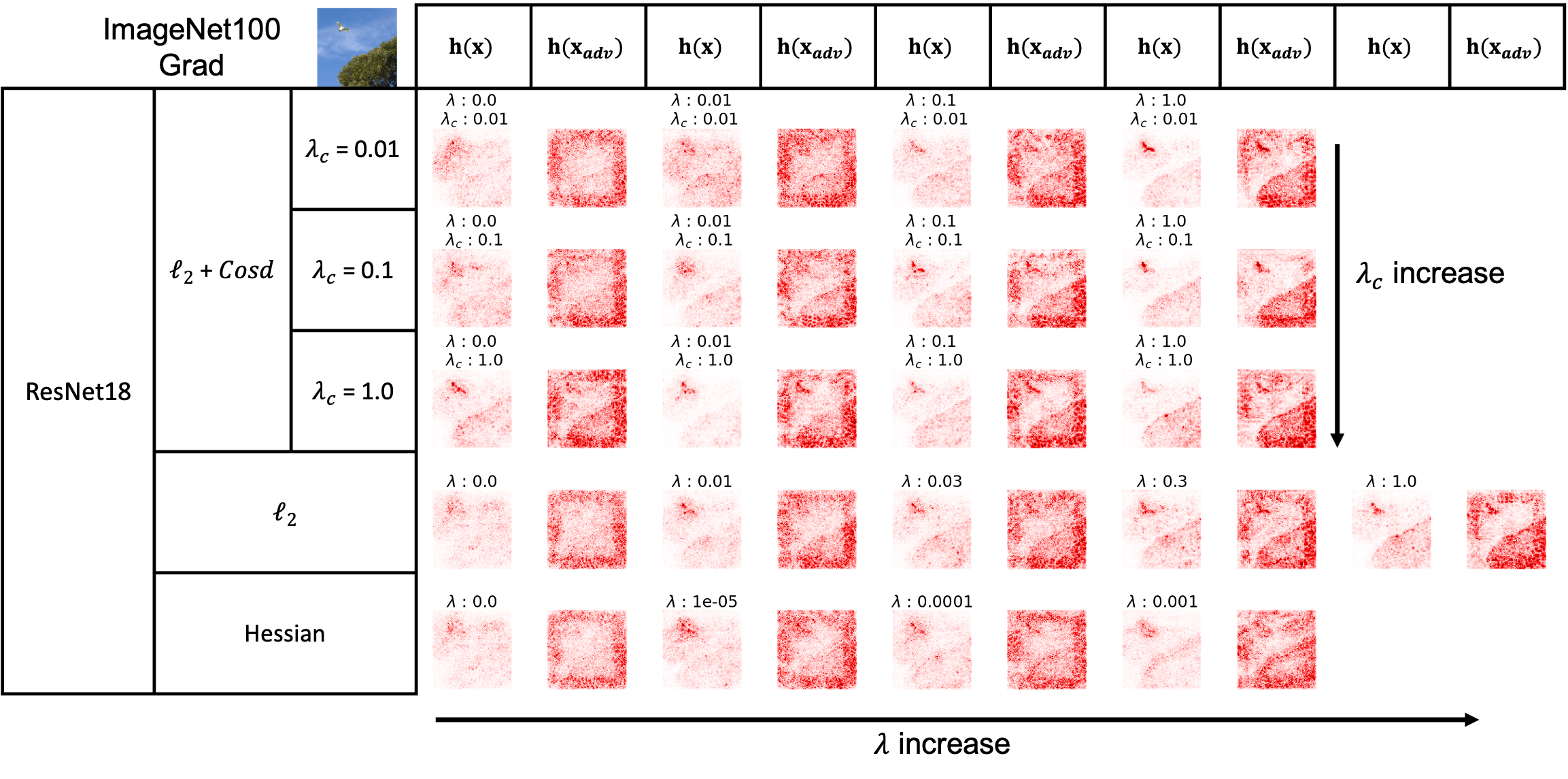}}
\caption{A visualization of the original and manipulated Gradient attribution for a ImageNet100 image with each model. The CE model corresponds to the $\ell_2$ and Hessian with $\lambda=0$. The AAM has been performed by PGD$_{\ell_\infty}(\epsilon=2)$}
\label{fig:imagenet100_img1}
\end{center}
\vskip -0.2in
\end{figure}

\begin{figure}[ht]
\vskip 0.2in
\begin{center}
\centerline{\includegraphics[width=\columnwidth]{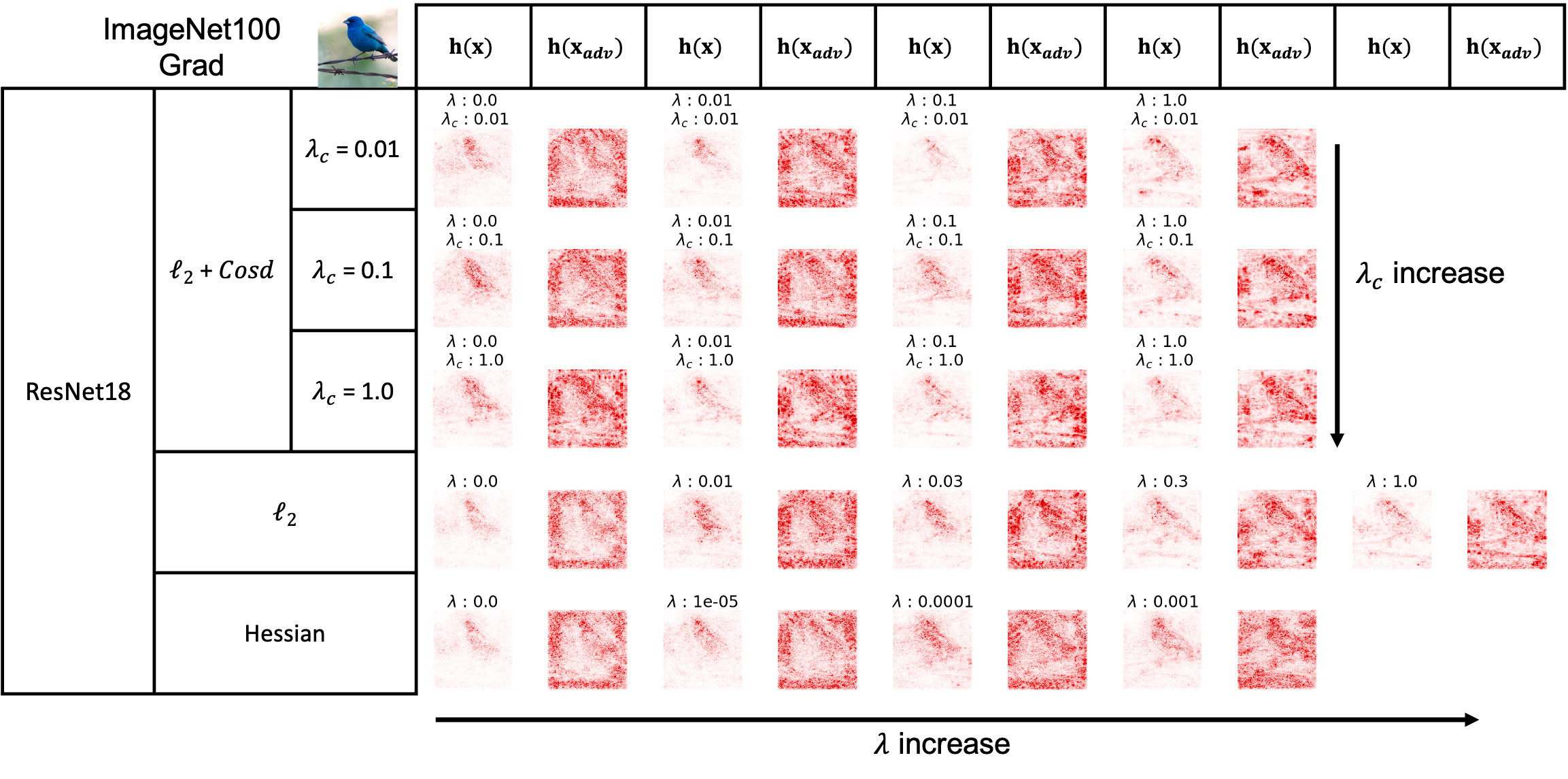}}
\caption{Visualization of the original and manipulated Gradient attribution for a Imagenet100 image with each model. The CE model corresponds to the $\ell_2$ and Hessian with $\lambda=0$. The AAM has been performed by PGD$_{\ell_\infty}(\epsilon=2)$}
\label{fig:imagenet100_img2}
\end{center}
\vskip -0.2in
\end{figure}

\begin{figure}[ht]
\vskip 0.2in
\begin{center}
\centerline{
 \captionsetup{width=0.8\textwidth}
 \includegraphics[width=0.8\textwidth]{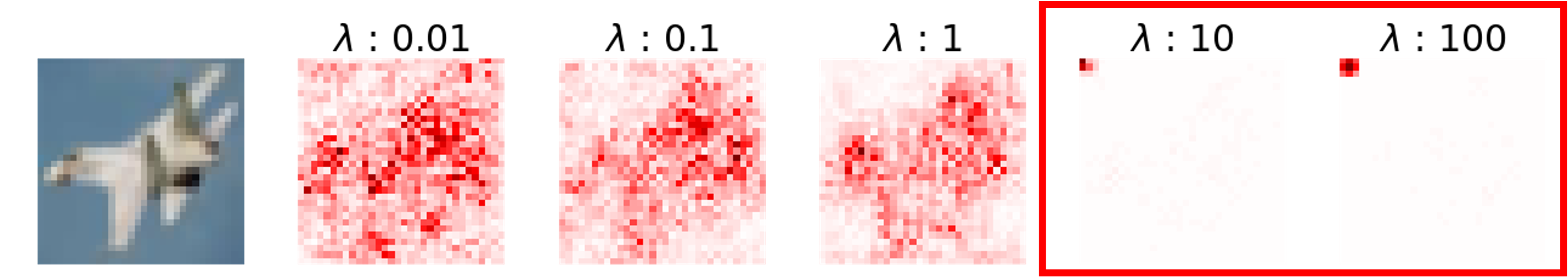}}
\caption{Visualization of the only Cosd robust criterion. Large coefficients of Cosd robust criterion make the ruined attribution maps in $\lambda = 10, 100$}
\label{fig:cossim4_ruin}
\end{center}
\vskip -0.2in
\end{figure}

In Figure \ref{fig:vis_all}, we visualized feature attribution maps of the selected models. In Figure \ref{fig:cifar10_img1} and \ref{fig:cifar10_img2}, we visualized feature attribution maps for CIFAR10 dataset with each model architecture and regularization methods, with varying the hyperparameter $\lambda$. We can observe that the attribution map is more clear when the $\lambda$ increases. We also included the visualization of every trained model with ImageNet100 dataset in Figure \ref{fig:imagenet100_img1} and \ref{fig:imagenet100_img2}.
An interesting result is given in Figure \ref{fig:cossim4_ruin} that regularizing a model with only cosine similarity and high regularization constant ruins the attribution. This result encourages us to use both $\ell_2$ and cosine distance as regularizers. 




\begin{sidewaystable}[ht]%
\centering
\caption{RPS of Grad and xGrad for every regularization method and hyperparameters.}
\label{tab:appendix_rps1}
\resizebox{\columnwidth}{!}{
\begin{tabular}{ccc|ccc|ccc|ccc|ccc|ccc|ccc|ccc}
\toprule
            &          &             &          &      &          & \multicolumn{9}{c}{Grad} & \multicolumn{9}{c}{xGrad} \\
            &          &             &          &      &          & \multicolumn{3}{c}{RPS($\epsilon=$4.0)} & \multicolumn{3}{c}{RPS($\epsilon=$8.0)} & \multicolumn{3}{c}{RPS($\epsilon=$16.0)} & \multicolumn{3}{c}{RPS($\epsilon=$4.0)} & \multicolumn{3}{c}{RPS($\epsilon=$8.0)} & \multicolumn{3}{c}{RPS($\epsilon=$16.0)} \\
            &          &             &          &      &  Accuracy &              Cossim &          PCC &         SSIM &              Cossim &          PCC &         SSIM &               Cossim &          PCC &         SSIM &              Cossim &          PCC &         SSIM &              Cossim &          PCC &         SSIM &               Cossim &          PCC &         SSIM \\
Dataset & Model & Regularizer & $\lambda$ & $\lambda_c$ &           &                     &              &              &                     &              &              &                      &              &              &                     &              &              &                     &              &              &                      &              &              \\
\midrule
\multirow{59}{*}{CIFAR10} & \multirow{36}{*}{LeNet} & \multirow{8}{*}{ATEX \cite{tang2022defense}} & \multirow{2}{*}{0.100000} & 2.00 &  87.6±0.3 &         0.925±0.004 &  0.823±0.007 &  0.801±0.007 &         0.850±0.007 &  0.648±0.010 &  0.636±0.009 &          0.758±0.007 &  0.425±0.008 &  0.438±0.008 &         0.916±0.005 &  0.842±0.006 &  0.853±0.004 &         0.840±0.007 &  0.701±0.009 &  0.733±0.007 &          0.748±0.008 &  0.527±0.010 &  0.587±0.010 \\
\bottomrule
\end{tabular}

}
\end{sidewaystable}

\begin{sidewaystable}[ht]%
\centering
\caption{RPS of GBP and LRP for every regularization method and hyperparameters.}
\label{tab:appendix_rps2}
\resizebox{\columnwidth}{!}{
\begin{tabular}{ccc|ccc|ccc|ccc|ccc|ccc|ccc|ccc}
\toprule
            &          &             &          &      &          & \multicolumn{9}{c}{GBP} & \multicolumn{9}{c}{LRP} \\
            &          &             &          &      &          & \multicolumn{3}{c}{RPS($\epsilon=$4.0)} & \multicolumn{3}{c}{RPS($\epsilon=$8.0)} & \multicolumn{3}{c}{RPS($\epsilon=$16.0)} & \multicolumn{3}{c}{RPS($\epsilon=$4.0)} & \multicolumn{3}{c}{RPS($\epsilon=$8.0)} & \multicolumn{3}{c}{RPS($\epsilon=$16.0)} \\
            &          &             &          &      &  Accuracy &              Cossim &          PCC &         SSIM &              Cossim &          PCC &         SSIM &               Cossim &          PCC &         SSIM &              Cossim &          PCC &         SSIM &              Cossim &          PCC &         SSIM &               Cossim &          PCC &         SSIM \\
Dataset & Model & Regularizer & $\lambda$ & $\lambda_c$ &           &                     &              &              &                     &              &              &                      &              &              &                     &              &              &                     &              &              &                      &              &              \\
\midrule
\multirow{59}{*}{CIFAR10} & \multirow{36}{*}{LeNet} & \multirow{8}{*}{ATEX \cite{tang2022defense}} & \multirow{2}{*}{0.100000} & 2.00 &  87.6±0.3 &         0.983±0.002 &  0.966±0.006 &  0.958±0.007 &         0.960±0.003 &  0.921±0.010 &  0.907±0.011 &          0.901±0.008 &  0.802±0.020 &  0.788±0.014 &         0.998±0.001 &  0.994±0.001 &  0.992±0.001 &         0.995±0.000 &  0.982±0.001 &  0.976±0.002 &          0.981±0.001 &  0.938±0.004 &  0.919±0.005 \\
\bottomrule
\end{tabular}
}
\end{sidewaystable}

\begin{sidewaystable}[ht]%
\centering
\caption{Insertion and Adv-Insertion score of Grad, xGrad, GBP, and LRP for every regularization method and hyperparameters.}
\label{tab:appendix_insertion}
\resizebox{0.7\columnwidth}{!}{
\begin{tabular}{ccc|ccc|cc|cc|cc|cc}
\toprule
            &          &             &          &      & \multicolumn{9}{c}{ } \\
            &          &             &          &      &          & \multicolumn{2}{c}{Grad} & \multicolumn{2}{c}{xGrad} & \multicolumn{2}{c}{GBP} & \multicolumn{2}{c}{LRP} \\
            &          &             &          &      &  Accuracy &    Insertion & Adv-Insertion &    Insertion & Adv-Insertion &    Insertion & Adv-Insertion &    Insertion & Adv-Insertion \\
Dataset & Model & Regularizer & $\lambda$ & $\lambda_c$ &           &              &               &              &               &              &               &              &               \\
\midrule

\multirow{59}{*}{CIFAR10} & \multirow{36}{*}{LeNet} & \multirow{8}{*}{ATEX \cite{tang2022defense}} & \multirow{2}{*}{0.100000} & 2.00 &  87.6±0.3 &  0.338±0.010 &   0.285±0.007 &  0.420±0.008 &   0.368±0.009 &  0.486±0.036 &   0.404±0.028 &  0.569±0.012 &   0.488±0.012 \\
            &          &             &          & 4.00 &  87.8±0.2 &  0.331±0.001 &   0.279±0.006 &  0.413±0.001 &   0.364±0.003 &  0.492±0.023 &   0.409±0.014 &  0.570±0.012 &   0.485±0.017 \\
\cline{4-14}
            &          &             & \multirow{2}{*}{0.300000} & 2.00 &  87.9±0.5 &  0.350±0.010 &   0.292±0.007 &  0.429±0.007 &   0.376±0.006 &  0.514±0.012 &   0.430±0.010 &  0.573±0.007 &   0.493±0.006 \\
            &          &             &          & 4.00 &  87.8±0.2 &  0.340±0.012 &   0.283±0.006 &  0.421±0.011 &   0.366±0.007 &  0.515±0.018 &   0.430±0.017 &  0.575±0.008 &   0.491±0.009 \\
\bottomrule
\end{tabular}
}
\end{sidewaystable}

\begin{table}[ht]%
\centering
\caption{Adv-Insertion score of Grad, xGrad, GBP, and LRP for selected model in $\epsilon = 2,4$}
\label{tab:appendix_adv_ins}
\resizebox{\columnwidth}{!}{
\begin{tabular}{c|c|cc|cc|cc|cc}
\toprule
\multirow{2}{*}{\shortstack{Dataset\\Model}} & \multirow{2}{*}{Regularizer} & \multicolumn{2}{c|}{Grad \shortcite{simonyan2014deep}} & \multicolumn{2}{c|}{xGrad \shortcite{shrikumar2017learning}} & \multicolumn{2}{c|}{GBP \shortcite{springenberg2014striving}} & \multicolumn{2}{c}{LRP \shortcite{bach2015pixel}} \\
             &     &             $\epsilon =2$ & $\epsilon =4$ &            $\epsilon =2$ & $\epsilon =4$ &    $\epsilon =2$           & $\epsilon =4$ &        $\epsilon =2$ & $\epsilon =4$ \\
\midrule
\multirow{6}{*}{\shortstack{CIFAR10\\LeNet}} & CE only &            30.2 &  29.1 &              39.1 &  37.4 &            46.4 &  43.0 &         53.7 &  48.4 \\
             & ATEX \cite{tang2022defense} &     30.4 &  28.9 &             39.4 &  37.5 &             46.8 &  42.7 &         55.2 &  49.9 \\
             & IGA \cite{singh2020attributional} &             35.0 &  34.0 &                               45.6 &  44.6 &                         56.1 &  54.7 &                         57.6 &  55.1 \\
             & Hessian \cite{dombrowski2022towards} &            33.9 &  31.7 &               43.6 &  41.6 &            57.4 &  54.8 &         58.3 &  56.3 \\
             & $\ell_2$      &  37.0 &  33.4 &           45.2 &  42.3 &          57.7 &  55.4 &         58.7 &  56.5 \\
             & $\ell_2$+ Cosd (ours) &        39.3 &  35.8 &         47.9 &  44.8 &              59.7&  57.4 &        59.2 &  57.1 \\
\bottomrule
\end{tabular}
}
\end{table}
\newpage
\clearpage

\bibliography{custom, bibfile}